%% file: main.tex
\documentclass[twoside,11pt]{article}

\usepackage{amsthm}
\usepackage{dmlr2e}

\usepackage{graphicx}%
\usepackage{multirow}%
\usepackage{amsmath,amssymb,amsfonts}%
\usepackage{mathrsfs}%
\usepackage[title]{appendix}%
\usepackage{xcolor}%
\usepackage{textcomp}%
\usepackage{manyfoot}%
\usepackage{booktabs}%
\usepackage{algorithm}%
\usepackage{algorithmicx}%
\usepackage{algpseudocode}%
\usepackage{listings}%

\usepackage[utf8]{inputenc} 
\usepackage[T1]{fontenc}    
\usepackage{hyperref}       
\usepackage{url}            
\usepackage{booktabs}       
\usepackage{amsfonts}       
\usepackage{nicefrac}       
\usepackage{microtype}      
\usepackage{xcolor}         

\usepackage{tablefootnote}
\usepackage{caption}
\usepackage{subcaption}
\usepackage{physics}
\usepackage{amsmath}
\usepackage{tikz}
\usepackage{mathdots}
\usepackage{yhmath}
\usepackage{cancel}
\usepackage{color}
\usepackage{siunitx}
\usepackage{array}
\usepackage{multirow}
\usepackage{amssymb}
\usepackage{amsmath}
\usepackage{gensymb}
\usepackage{tabularx}
\usepackage{extarrows}
\usepackage{booktabs}
\usetikzlibrary{fadings}
\usetikzlibrary{patterns}
\usetikzlibrary{shadows.blur}
\usetikzlibrary{shapes}
\usepackage{pifont}
\usepackage{wrapfig}
\usepackage{marvosym}

\usepackage{lastpage}
\dmlrheading{12}{2025}{1-\pageref{LastPage}}{12/25; Revised 6/26}{--/--}{21-0000}{Paul Kahlmeyer, Henrik Voigt, Michael Habeck and Joachim Giesen} 

\def\openreview{\url{}} 

\ShortHeadings{ERBench}{Kahlmeyer et al.}
\firstpageno{1}

\begin{document}

\title{ERBench: A Benchmark and Testsuite for Equation Discovery Algorithms}

\author{%
  \name Paul Kahlmeyer\thanks{These authors contributed equally to this work.} \email paul.kahlmeyer@uni-jena.de \\
  \name Henrik Voigt\footnotemark[1] \email henrik.voigt@uni-jena.de \\
  \name Michael Habeck\footnotemark[1] \email michael.habeck@uni-jena.de \\
  \name Joachim Giesen\footnotemark[1] \email joachim.giesen@uni-jena.de \\
  \addr Department of Computer Science\\
  University of Jena\\
  Inselplatz 5, 07743 Jena, Germany
}

\editor{} 

\maketitle

\begin{abstract}
Equation discovery aims to automate the discovery of scientific models in the form of mathematical equations from data. 
Technically, equation discovery is implemented by symbolic regression algorithms.
Performance of symbolic regression for equation discovery is measured along two dimensions: Prediction accuracy on test data, and recovery of known groundtruth formulas.
For standard regression, accuracy is typically measured on in-domain test data, for instance, by splitting a data set randomly into training and test data.
While this makes sense for in-domain interpolation, which is the common goal in ordinary regression, it can be a misleading proxy for true model discovery and generalization. 
The obvious alternative is to measure out-of-domain accuracy.
However, obtaining challenging out-of-domain test data is a non-trivial problem.
Therefore, we focus on equation recovery for evaluating symbolic regression algorithms for equation discovery.
The rationale is that symbolic regression algorithms that perform well in recovering known groundtruth formulas are good candidates to perform well in unknown equation discovery.
Existing benchmarks for symbolic regression include equation recovery tasks, however, with only a small number of groundtruth formulas that are publicly known.
Moreover, these benchmarks place less emphasis on evaluating the robustness of algorithms in terms of their behavior under changing dimensionality, sampling size, sampling distribution and sampling domain. 
This, however, is of central importance to practitioners wanting to discover equations for modeling natural phenomena, since data is almost certainly noisy and comes from diverse domains, distributions, and sample sizes.
To fill this gap, we introduce the Equation Recovery Benchmark (ERBench), a new evaluation framework designed to rigorously assess algorithms explicitly targeting the task of equation discovery.
\end{abstract}

\vspace{1em}


\section{Introduction}
\label{sec:introduction}
The scientific method is fundamentally driven by the development of models that make bold, falsifiable predictions about the world. As Karl Popper argues, a theory's scientific merit lies not in its verification but in its capacity to be refuted through rigorous empirical testing~\citep{popper2005logic}. Confidence in a model grows when it survives demanding experiments designed to challenge it, particularly in novel, out-of-domain scenarios. 
In contrast, a model merely fitted to existing data in hindsight offers only retrodictive power in the form of post-hoc analysis but lacks the explanatory potential required for true scientific advancement.

\begin{figure}[t!]
  \centering
  \includegraphics[width=0.99\linewidth]{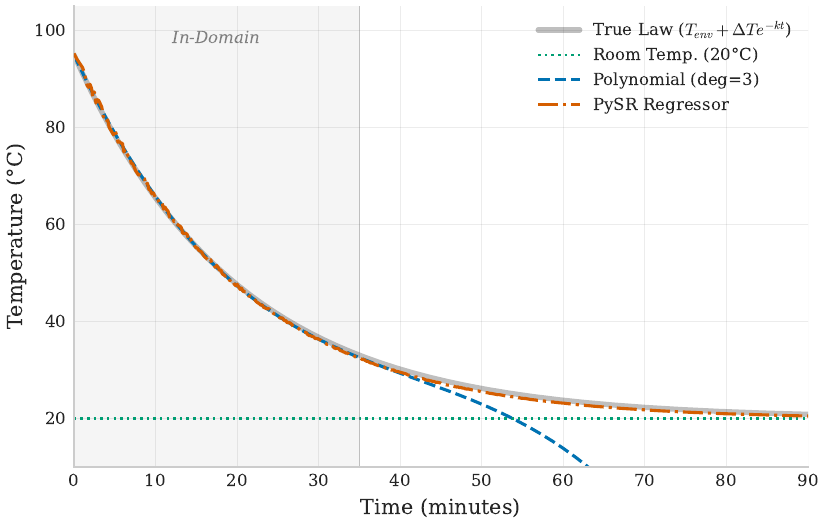}
    \caption{\textbf{Generalization ability of a symbolic equation compared to a polynomial regressor.} The figure contrasts the out-of-domain generalization capabilities of a proposed symbolic equation against a polynomial model. The ground truth is Newton's Law of Cooling, $T(t) = T_{\text{env}} + (T_0 - T_{\text{env}})e^{-kt}$, which describes the temperature of an object over time starting at temperature $T_0$ and cooling down to the temperature of the surrounding environment $T_{\text{env}}$ with rate $k$. A polynomial model is fitted to in-domain training data from the initial cooling phase (0–35 minutes), and the best-fitting polynomial is selected using the BIC criterion. While the polynomial achieves high accuracy on the training domain, it fails to generalize and makes physically implausible predictions for the out-of-domain samples (durations longer than 35 minutes going below room temperature). In contrast, the symbolic equation proposed by the state-of-the-art symbolic regression algorithm, PySR~\citep{pysr_cranmer23}, generalizes more robustly.} 
  \label{fig:generalization_example}
\end{figure}

\noindent However, Popper's ideal does not encompass the primary goal of all modeling endeavors. In fields like engineering and applied sciences, the objective is often to create models with high predictive accuracy within a well-defined domain. The goal is not to make bold extrapolations, but to achieve reliable interpolation. Models such as polynomial regressors, neural networks~\citep{rumelhart1985learning}, and Gaussian process regressors~\citep{rasmussen2003gaussian} excel at this task. While their predictive power on in-domain data can be substantial, they are often treated as "black boxes" with limited explanatory power and typically fail when tasked with making predictions outside their training distribution (see Figure~\ref{fig:generalization_example}).

\noindent In contrast, the ambition of scientific modeling is the pursuit of generalization and explanation. The goal is to discover models that are not only predictive, but also provide insight into the underlying mechanisms of a system. Such models must be capable of extrapolation. That means making falsifiable predictions under novel conditions, which is central to the scientific method. 

\noindent The goal of equation discovery is to support scientists in the process of making new scientific discoveries, that is, finding equations that explain phenomena in nature. By searching for parsimonious, symbolic mathematical expressions that describe data, equation discovery aims to uncover the fundamental principles governing a system, yielding models that are inherently generalizable, interpretable, and falsifiable.


\noindent While the philosophical goal of scientific modeling is clear, translating this ideal into a computational process is non-trivial. 
In the restricted scenario of equation discovery, this is achieved through symbolic regression algorithms. Symbolic regression algorithms are trained on a labeled data set from which a symbolic regression function is computed.
So far, there is no consensus on the evaluation of symbolic regression algorithms for equation discovery, primarily because symbolic regression algorithms are also used for standard regression, that is, creating models with high predictive accuracy, however, only within a well-defined training domain. Following Popper's framework, the aim of scientific equation discovery is generalization and explanation, which is not adequately evaluated by prediction accuracy on the training domain. Symbolic regression algorithms should be evaluated by their out-of-domain generalization ability. 
However, testing symbolic regression algorithms on out-of-domain test data can be challenging, because the test data can be insufficient to discriminate between models.
For instance, the formulas $x^2$ and $x^2+\log x$ can not be discriminated for large values of $x$ and slightly noisy data. Picking up the log-term, however, can be scientifically significant. 

\noindent A strong measure of out-of-domain generalizability is the ability to recover known ground truth formulas from training data. Equation recovery can be seen as an analog to equation discovery, except that, in this setting, the ground-truth formulas are known. Algorithms that excel at recovering known models from data are strong contenders for discovering new equations from observed data, thus supporting researchers in their scientific work.\\ 

\noindent In this work, we present the \textit{Equation Recovery Benchmark (ERBench)}, an evaluation framework designed to assess the performance of symbolic regression algorithms on the task of equation recovery. The benchmark implements the ideas of a \emph{common task framework}~\citep{Liberman10} that is characterized by a public set of well-defined tasks and criteria (developed in consultation with researchers) and secret evaluation tasks that are withheld for periodic public evaluations. In ERBench the task is the recovery of known ground truth formulas and the criterion is a high recovery rate. The contributions of our evaluation framework are summarized as follows: 

\begin{itemize}

    \item \textbf{Generalization:} In machine learning, generalization is typically measured by a model's predictive accuracy on unseen data. However, for the specific task of symbolic regression, we propose a stricter and more meaningful standard. While approximation is sufficient for engineering, we posit that for scientific applications a more adequate measure of success is a symbolic regression algorithm's ability to recover the ground-truth symbolic equation from which the data have been generated. An algorithm capable of this has generalized beyond the data to the underlying law of the system. We argue that an algorithm that has demonstrated to reliably rediscover known laws serves as a credible tool for discovering novel ones. Consequently, our benchmark prioritizes evaluation metrics that measure symbolic equivalence with the ground truth, rather than relying solely on predictive error.

    \item \textbf{Robustness:} We provide training datasets of various sizes, distributions and domains to assess the robustness of equation discovery algorithms. This allows practitioners to clearly understand an algorithm's capabilities and limitations in practical applications.

    \item \textbf{Dataset:} The public training set compiles a wide range of ground truth formulas from different scientific domains and a large, synthetic dataset of formulas of varying difficulty. 
    The secret test set is generated by a parameterized, reproducible procedure that ensures fairness and reliable results while minimizing the likelihood that items appear in pretraining corpora, which is essential for fairly evaluating modern methods such as LLMs.

    \item \textbf{Leaderboard:} We provide an accurate picture of the current state of the art at any given time by hosting an ongoing competition that is both easily accessible to researchers and straightforward to maintain.
\end{itemize}

\noindent This paper is structured as follows: First, we discuss the landscape of existing benchmarks in symbolic regression and equation discovery and their specific focus areas. Then, we review state-of-the-art methods in equation discovery, analyzing the respective strengths and weaknesses of different paradigms. Taking these into account, we introduce the Equation Recovery Benchmark, detailing its composition, design principles, and evaluation metrics. To provide a baseline for future work, we conduct an evaluation of state-of-the-art algorithms on the secret test set of the benchmark. We conclude by discussing these results to identify key challenges and outline promising directions for future research.

\section{Related Work}
\label{sec:related_work}
Our work builds upon a rich body of work to benchmark symbolic regression and automated scientific discovery algorithms. To contextualize our contribution, we first survey the evolution of symbolic regression benchmarks. We then broaden our scope to discuss ambitious challenges in automating the full scientific discovery process. 

\subsection{Symbolic Regression Benchmarks} 
Due to the lack of a common benchmark, early developers of symbolic regression algorithms evaluated their algorithms on their own ground truth formulas~\citep{uy2011semantically, keijzer2003}. 
Consequently, a comparative overview of the performance of different symbolic regression algorithms was absent.
This issue was first highlighted by \citet{benchmark_mcdermott12}, who compiled a collection of synthetic formulas from the symbolic regression literature, but left the task of benchmarking to future research. 
A first substantial attempt at benchmarking symbolic regression algorithms was done by \citet{benchmark_lacava18} on the Penn Machine Learning Benchmark (PMLB) collection of regression problems~\citep{PMLB_17}. 
The comparison uses standard regression metrics like predictive accuracy on in-domain data and is restricted to genetic programming algorithms.
\citet{feynmanAI_udrescu20} introduced the Feynman dataset as a collection of $119$ physics formulas from the Feynman lectures to benchmark their algorithm. In follow-up work, \citet{srbench_lacava21} widened the scope of the benchmark to a broader range of symbolic regression algorithms. 
Their benchmark results, dataset and evaluation code was made publicly available under the name SRBench. 
With the publication of the evaluation dataset, however, new symbolic regression algorithms could no longer be compared fairly.
This issue was addressed by the \textit{Interpretable Symbolic Regression for Data Science} competitions, hosted at the genetic and evolutionary computation conferences GECCO'22~\citep{competition_lacava22} and GECCO'23~\citep{competition_defranca23}.
Researchers participating in the competitions were encouraged to train their algorithms on SRBench. 
The algorithms are then evaluated on the known but also on new and thus unknown formulas. In its most recent iteration~\citep{Imaildeia25}, SRBench 2.0 focuses on ranking evaluated algorithms by predictive accuracy, following the ideas of classical regression methods. 

\noindent Proposed extensions to SRBench have been made in the form of SRDS~\citep{matsubara2024srds}, which attempt to align the numerical constants in formulas and sampling domains used in SRBench with values more common in real world physics. LLM-SRBench~\citep{shojaee2025llm} is designed to mitigate the risk of training data leakage when evaluating LLMs by employing novel and adapted equations to create a benchmark that is less likely to have been seen during pre-training. The paper SRBench++~\citep{de2024srbench++} introduces expert evaluations to rate the interpretability of models on domain-specific tasks. Recently, in their influential work, PySR~\citep{pysr_cranmer23},  the authors introduce EmpiricalBench, a set of real physics problems that have been discovered by scientists from noisy and imperfect data. They argue that an algorithm should be considered useful for scientific equation discovery if it can discover relations of this kind. 
Our work builds upon this by systematizing the ideas into a comprehensive evaluation framework comprising a large, diverse set of public formulas, combined with a secret test set involving rigorous evaluation across variations in sample size, domain and distribution. We believe that this will facilitate the development of robust algorithms to help researchers discover new scientific equations.

\subsection{Equation Discovery Benchmarks}
Recent work highlights a distinction between equation discovery from static datasets and the broader challenge of autonomous scientific experimentation~\citep{kramer2023automated}. To bridge this gap, a new class of research has emerged, focusing on creating AI scientist environments that automate the entire closed-loop scientific process—from hypothesis generation to experimentation and analysis. This ambitious goal is aligned with grand challenges like the Nobel Turing Challenge, which aims to develop AIs capable of major autonomous discoveries~\citep{kitano2021nobel}. In these systems, equation discovery serves as one crucial module within a larger agent architecture. Prominent examples include Science-Gym~\citep{cerrato2024science}, which requires agents to actively design experiments to discover phenomena, and PhysGym~\citep{chen2025physgym}, which evaluates an agent's ability to formulate and update hypotheses based on experimental outcomes. The primary focus of these environments is to evaluate the agent's overall performance across the entire scientific workflow. Our work is intentionally distinct from this line of research. By design, ERBench isolates the equation discovery task from the complexities of active data acquisition and experimental design. This focused approach is complementary: we provide a static but comprehensive testbed to enable a deep, fine-grained evaluation of an algorithm's core ability to recover ground-truth equations from a given dataset. We therefore do not measure the preceding steps of hypothesis generation or data gathering, but instead provide a rigorous benchmark for the pivotal analysis stage. 

\section{State of the Art in Equation Discovery}
\label{sec:review}
Equation discovery is understood as a specialized sub-field of symbolic regression, distinguished by its ultimate objective. 
The primary goal of symbolic regression is to find a symbolic model that minimizes a predictive error metric, such as mean squared error, on a given dataset. In this respect, its goal is analogous to classical regression methods like polynomial regression, with the crucial distinction that symbolic regression searches for both the functional structure and its parameters, whereas classical methods operate on a predefined structure. 
By contrast, equation discovery seeks to recover the underlying generative equation itself, emphasizing parsimony and symbolic equivalence rather than predictive accuracy alone.

\paragraph{Symbolic Regression for Equation Discovery.} 
Technically, equation discovery can be conducted using symbolic regression algorithms. These algorithms aim at finding a function that can be constructed from a set of elementary operators, such as $+,-,*,/,\log, \exp, \sin$ and $\cos$.
Classical regression expresses functions as linear combinations of basis functions, reducing training to parameter optimization, typically solvable by gradient-based methods.
Equation discovery does not reduce to parameter optimization, but also includes an \emph{architecture-search} step, that is, a search over expression trees or expression DAGs. 
The nodes of the trees or DAGs are annotated by elementary operators.
Equation discovery methods typically need fewer trainable parameters to achieve the same flexibility as conventional regression methods.\\

\begin{figure}[t!]
    \centering
    \resizebox{\textwidth}{!}{
        \input{figures/taxonomy}
    }
    \caption{
        \textbf{A taxonomy of symbolic regression workflows, distinguishing between pre-training-based and traditional approaches.} 
        The process can be divided into three distinct stages. 
        \textbf{(1) Pre-Training:} A phase unique to deep learning-based algorithms, where a general-purpose model learns underlying patterns from a vast corpus of equations (e.g., $x^2, \sin(x)$) and correspondingly sampled synthetic data.
        \textbf{(2) Training:} The classic equation discovery task. Given a single dataset of $m$ sample points $X\in\mathbb{R}^{m\times n}$ and target values $\boldsymbol{y}\in\mathbb{R}^m$, an algorithm (e.g., a genetic programming algorithm, sparse regressor, or a pre-trained algorithm) is used to find a specific symbolic function. Traditional methods begin at this stage.
        \textbf{(3) Inference:} The final, discovered regression function is used to make predictions on new, unseen data.
    }
    \label{fig:naming_convention}
\end{figure}
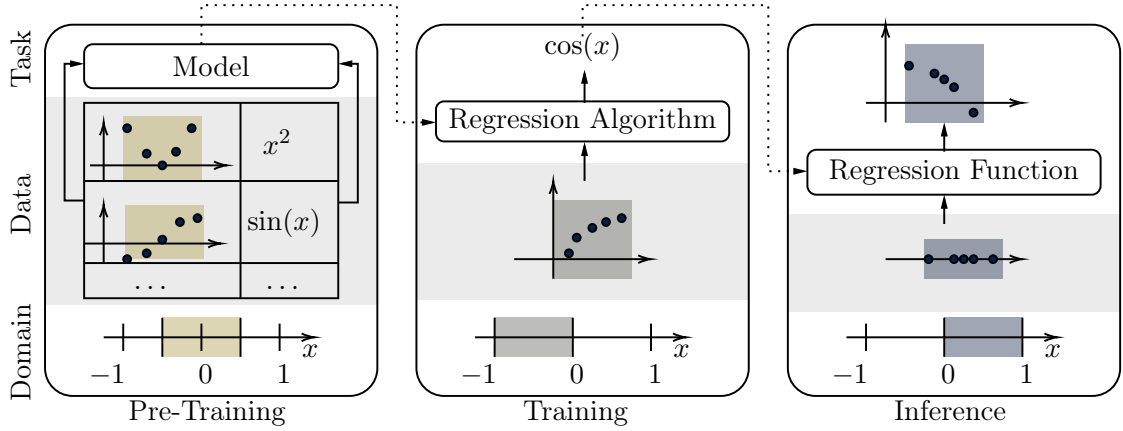

\noindent In contrast to classical regression algorithms that only employ a training phase to compute a formula, pre-training is used extensively in symbolic regression.
In our terminology that is illustrated in Figure~\ref{fig:naming_convention}, pre-training turns pre-training data into a symbolic regression algorithm that turns training data into a formula. 
The formula is used in inference to compute labels $y\in\mathbb{R}$ from features $x\in\mathbb{R}^n$. 
Pre-training data consist of formulas $f$ and feature-label pairs $\big(x,y=f(x)\big)$, where the features $x$ have been sampled from the domain of $f$.
Training data are feature-label pairs $(x,y)$.

\noindent Note, however, that not all symbolic regression algorithms are derived from a pre-training phase.
Therefore, we distinguish five categories of symbolic regression algorithms, namely, (1) extensions of conventional regression algorithms, (2) enumeration-based and (3) sampling-based search algorithms, (4) algorithms derived from a pre-training phase, and (5) hybrid algorithms that are derived from pre-training but also employ a search during the training phase. 
In the following, we briefly review algorithms from the five categories, describe their strengths and weaknesses. 
Crucially, we use this analysis to derive specific design principles for our benchmark, denoted throughout the text by the symbol \Forward.

\subsection{Extensions of Conventional Regression Methods}
By conventional regression, we refer to the approach of first selecting a finite set of basis functions, for instance, polynomials up to a certain degree, and then computing the coefficients for the basis functions by minimizing some loss function.
The space of all linear combinations of basis functions provides the hypothesis class for the regression approach. 
The prototypical example of a classical regression method is ordinary least squares regression where the basis functions are simply the coordinate functions $x_i$ of a feature vector $x\in\mathbb{R}^n$ and the least squares loss function minimizes the squared error between the predictions and observations on the training data.
The hypothesis class of ordinary least squares regression is the space of all linear functions on $\mathbb{R}^n$.

\paragraph{Sparse Regression.}
Sparse regression is a natural extension of conventional regression.
The key difference is that a sparsity inducing regularization term, typically an $\ell_1$-loss term, is added to the loss function.
The effect of the regularization term is setting many coefficients to zero.
Implementations of the sparse regression approach include fast function extraction (FFX) by~\citet{ffx_mcconaghy11} and sparse identification of nonlinear dynamics (SINDy) by~\citet{sparse_brunton16}.

\paragraph{Sparse Regression Networks.}
Sparse regression networks extend the idea of sparse regression to nested sums of basis functions, which, are called activation functions in this context.
These nested sums define a hierarchical network structure.
In the Equation Learner (EQL) network by~\citet{eql_lampert18}, a predefined library of functions such as, for instance, $\sin$, $\cos$, and $\log$ are used as activation functions, whereas Kolmogorov-Arnold Networks (KANs)~\citep{liu2025kan} use splines with trainable parameters.
Later, the splines in KANs can be replaced by symbolic formulas by using an additional symbolic regression step. 
Both approaches employ an $\ell_1$-regularization term in their loss function.\\ 

\noindent \Forward\, By design, sparse regression algorithms can only recover formulas from the hypothesis class that is given by the choice of basis functions. 
Sparse regression networks are limited in their expressiveness through the choice of the network architecture. Consequently, suitable recovery performance for these algorithms is only expected on the prescribed hypothesis classes or network architectures.
Therefore, a fair and comprehensive benchmark should provide a diverse collection of formulas from different hypothesis classes to verify robustness across structural biases.

\subsection{Enumeration-Based Search Algorithms}
\begin{figure}[t]
     \centering
     \begin{subfigure}[b]{0.45\textwidth}
         \centering
         \includegraphics[width=\textwidth]{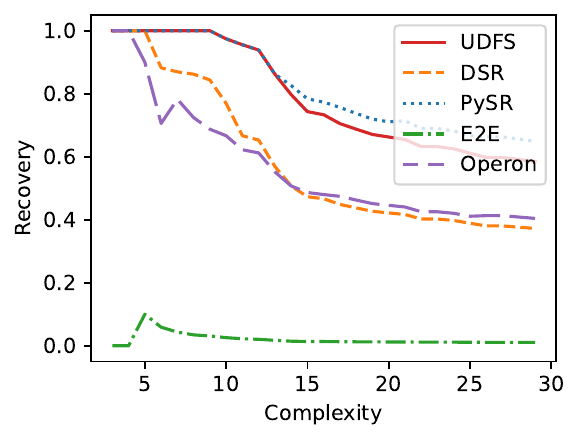}
     \end{subfigure}
     \hfill
     \begin{subfigure}[b]{0.45\textwidth}
         \centering
         \includegraphics[width=\textwidth]{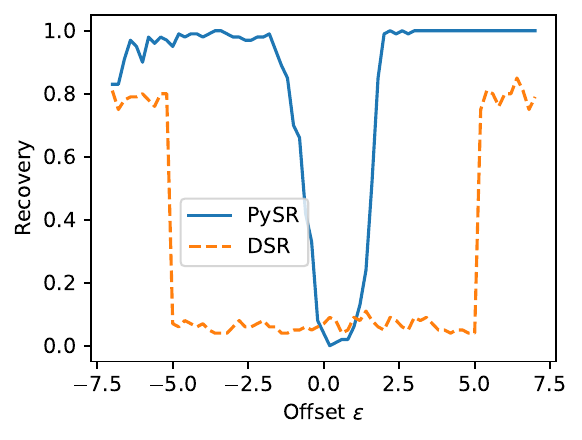}
     \end{subfigure}
    \caption{\textbf{Performance artifacts of state-of-the-art search-based algorithms.} Given samples from a ground-truth equation, we measure the percentage of correctly recovered equations. \textbf{Left:} The recovery performance of symbolic regression algorithms degrades when the representation complexity of the ground truth formulas increases. Shown here is the fraction of successfully recovered formulas from the Feynman equations dataset~\citep{feynmanAI_udrescu20} whose expression complexity, measured by the number of nodes in a minimal expression tree, is upper bounded by the values on the x-axis.
    \textbf{Right:} The recovery performance of distribution based search algorithms depends on the sampling domain of the training data. Shown are recovery rates for two distribution based search algorithms on 100 equally spaced training data points of the function $f(x_0) = \exp(x_0) - x_0^3$ in the interval $[-5+\varepsilon, 5+\varepsilon]$. Recovery rates are averaged over 100 runs.}
    \label{fig:related_work_search_based}
\end{figure}

Search-based symbolic regression algorithms represent formulas as expression trees, DAGs, or strings, and explore these spaces to find equations that fit the data. A straightforward implementation of the search-based paradigm exhaustively enumerates the space of all expressions with small representations. 
If the true expression has a small representation, then it will be recovered by an enumeration-based search algorithm, independently of the sampling domain.
Since the search spaces grow at least exponentially in the number of operators, the straightforward implementation of the enumeration-based search paradigm is restricted to impractically small expressions~\citep{esr_bartlett23}.
However, as can be seen in Figure~\ref{fig:related_work_search_based} (left) recovery with increasing complexity is not only an issue for enumeration-based search approaches, but for all known symbolic regression algorithms. 
There are, however, different ideas and techniques to scale up enumeration-based search.
\citet{gpusearch_ruan24, muthyala2025symanticefficientsymbolicregression} make use of the massive parallelism provided by GPUs to speed up the search.
The AIFeynman symbolic regression algorithm~\citep{feynmanAI_udrescu20, feynmanAI2_udrescu20} iteratively cuts down the search space by exploiting statistical properties such as symmetries.
\citet{substitutions_kahlmeyer25} search for valid substitutions to reduce the number of variables and thus the size of the search space.
Sometimes, however, it can be beneficial to also add new variables that represent small expressions in the original variables~\citep{udfs_kahlmeyer24}.
Such variable augmentations can reduce the number of operators and thus the overall size of an expression.\\

\noindent \Forward\, Since enumeration-based search approaches, but also other approaches, are limited to expressions with small representations, it is important to map out the representation complexity boundary (e.g. the number of operators and the dimensionality) up to which expressions can be reliably recovered. 
Therefore, a benchmark should cover a wide range of representation complexities.

\subsection{Sampling-Based Search Algorithms}
The majority of search-based symbolic regression algorithms try to avoid an exhaustive enumeration by sampling from a distribution over equations.
These sampling based algorithms differ in the way they sample and refine the underlying distribution.
A unifying feature of all current sampling-based search algorithms is that the distribution updates are guided by numerical closeness to the training data: encountered equations with low error (or high likelihood/reward) become more probable. Thus, if a representation, for instance an expression tree, of a formula that achieves high accuracy on the training data has been found, then these algorithms search in the neighborhood of this representation. Since the numerical accuracy depends on the samples themselves, the performance of distribution-based search algorithms will, in contrast to enumeration-based search algorithms, depend on the sampling domain. An example is shown in Figure~\ref{fig:related_work_search_based} (right). This broad category encompasses several distinct algorithmic paradigms, differentiating primarily in how they represent and update the distribution over expressions.

\paragraph{Genetic Programming.} 
The most prominent class of algorithms in this category are genetic programming algorithms that were first introduced by~\citet{koza1994genetic}. 
Equations are typically represented by expression trees. 
Starting from an initial population of expression trees, the best performing expression trees of the current population are recombined by (ex-)changing subtrees into the next population.
In genetic programming, the distribution over equations is only defined \emph{implicitly} via the current population of expressions. 
The algorithm specific recombination rules then define the adjustment of this distribution.
Prominent implementations of the genetic programming paradigm are Eureqa~\citep{eureqa_lipson09}, gplearn~\citep{gplearn_stephens16}, and PySR~\citep{pysr_cranmer23}.
In contrast to enumeration-based search algorithms, the performance of genetic programming algorithms seems to depend more on the size of the training sample, as can be seen in Table~\ref{tab:robustness_num_samples}.
\begin{table}[t!]
    \small
    \caption{\textbf{Evaluating algorithms under sparse data conditions.} Comparison of the training data efficiency of the enumeration-based symbolic regression algorithm UDFS~\citep{udfs_kahlmeyer24} and the genetic programming algorithm PySR~\citep{pysr_cranmer23}. Shown is the average rate of recovering a ground truth equation depending on the number of training data points per regression problem. The reported recovery rates are measured on 40 equations from the Feynman equations dataset~\citep{feynmanAI_udrescu20}. We report averages and standard deviations over five training data sets for each training data set size.\\} 
    \centering
    \begin{tabular}{lccccc}
    Samples per dimension&5&10&20&40&80\\
    \midrule
    UDFS&$1.0\pm0.0$&$1.0\pm0.0$&$1.0\pm0.0$&$1.0\pm0.0$&$1.0\pm0.0$\\
    PySR&$0.965\pm0.05$&$0.975\pm0.03$&$0.98\pm0.02$&$0.99\pm0.02$&$1.0\pm0.0$\\
    \end{tabular}
    \label{tab:robustness_num_samples}
\end{table}

\paragraph{Bayesian Approaches.}
A Bayesian approach to symbolic regression treats equation discovery as posterior inference, updating a prior distribution $p(\text{eq})$ into a training-data-dependent posterior distribution,
$$p(\text{eq}|X,y) \propto p(X, y|\text{eq})p(\text{eq})\,,$$
typically by Markov Chain Monte-Carlo (MCMC) methods. 
\citet{jin2020bayesian} model the prior sampling distribution by hand, whereas \citet{bayesianscientist_guimera20} learn it from data, more specifically from a corpus of closed-form mathematical formulas from Wikipedia.

\paragraph{Reinforcement Learning.}
Reinforcement learning formulates the sampling process as a sequential decision-making task, where an agent generates expressions step by step, typically in a predefined traversal order of the expression tree.
The distribution over expressions is \emph{explicitly} given and updated via some reward function, which depends on the numerical fit on the training data.
The first and prototypical example in this class is Deep Symbolic Regression (DSR)~\citep{dsr_petersen21}, which models the sampling distribution by a recurrent neural network (RNN). In an extension to DSR~\citep{dsr2_petersen21}, genetic programming is used to refine samples drawn from the distribution.
The Reinforcement Symbolic Regression Machine (RSRM)~\citep{xu2024rsrm} uses an interplay of Q-tables and Monte-Carlo Tree Search (MCTS) to model the distribution and also uses genetic programming for refinement.\\

\noindent \Forward\, Since the performance of sampling-based search algorithms depends on the domain and on the size of the training data, an equation discovery benchmark should provide various sampling domains and sample sizes.

\subsection{Pre-trained Symbolic Regression Algorithms}

\begin{figure}[t!]
     \centering
     \begin{subfigure}[b]{0.485\textwidth}
         \centering
         \includegraphics[width=\textwidth]{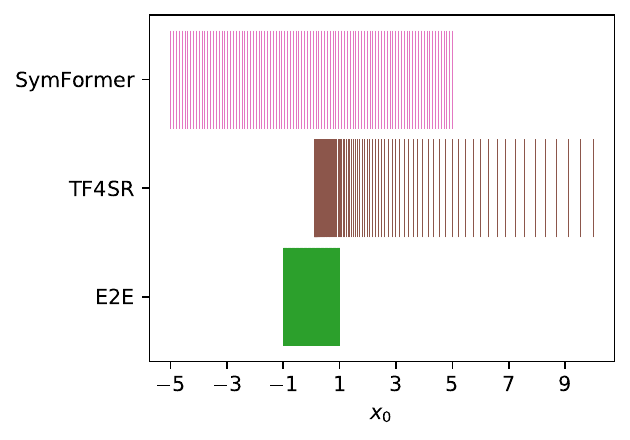}
     \end{subfigure}
     \hfill
     \begin{subfigure}[b]{0.45\textwidth}
         \centering
         \includegraphics[width=\textwidth]{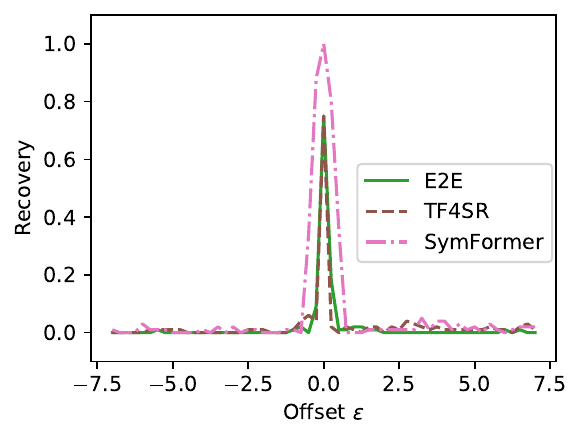}
     \end{subfigure}
    \caption{
    \textbf{The success of pre-training-based symbolic regression algorithms depends on the pre-training distribution.} \textbf{Left:} We transform 100 equally spaced points $x_0$ in $[0, 1]$, into pre-training data points by using the transformation $10\cdot x_0 - 5$ for SymFormer, $(x_0-\bar{x}_0)/\sigma_{x_0}$ for E2E, and $100^{x_0}/10$ for TF4SR. \textbf{Right:} 
    Percentage of correctly recovered ground truth for the three pre-training based algorithms on the function $f(x_0) = x_0^2$ when using the respective pre-training data points and a translational offset. For samples that span the interval $[a, b]$, the offset was scaled to $\varepsilon (b-a)$. Recovery rates are averaged over 100 runs. Performance degrades sharply out of the pre-training distribution. }
    \label{fig:related_work_pretraining}
\end{figure}

The training in enumeration- and sampling-based search is computationally demanding, because the size of the search space grows exponentially in the number of operators.
Pre-training-based symbolic regression algorithms move a significant part of the computational effort into a pre-training phase.
The pre-training data consists of a collection of formulas and samples from the domain of the formulas, together with the function values at the samples. 
The goal is to learn a function that maps labeled samples to formulas.
Training then simply becomes evaluating the latter function on the labeled samples that constitute the training data.\\

\noindent The majority of pre-trained symbolic regression algorithms pre-train a transformer that translates a sequence of sample points into a sequence of tokens that represents a formula. After the pretraining phase, the transformer thus falls into the category of a sampling-based search algorithm.
Concrete implementations differ in the way the sample points are encoded, in the representation of the formulas, in the handling of constants, and in the loss function.
Implementations of the pre-training paradigm for symbolic regression include neural symbolic regression that scales (NeSymReS)~\citep{biggio2021neural}, end-to-end symbolic regression with transformers (E2E)~\citep{kamienny22_transformer}, SymFormer~\citep{vastl22_symformer}, a transformer model for symbolic regression (TF4SR)~\citep{lalande23_transformer}, and symbolic regression as a multi-modal information fusion task
(MMSR)~\citep{mmsr_li25}. 
Most recent work focuses on enhancing the latent space~\citep{li2023transformerbased} or leveraging large scale LLMs~\citep{shojaee2025llmsr, merler_llm_2024}.
Another perspective on the problem of symbolic regression has been proposed by \citet{tian2025co}, who use a Q-network to pre-train a policy for reinforcement learning.
Since the main computational costs are shifted to the pre-training phase, pre-trained symbolic regression algorithms have a fast training phase. 
As we illustrate in Figure~\ref{fig:related_work_pretraining}, a common problem of pre-trained symbolic regression algorithms is their dependence on the domain from which the pre-training data are sampled.
If the training data are from a different domain than the pre-training data, then the recovery performance of pre-trained symbolic regression algorithms breaks down.\\

\noindent \Forward\, To reliably assess the recovery performance of pre-trained symbolic regression algorithms, it must be measured outside their pre-training domains. 
Therefore, an equation discovery benchmark should take samples from a wide variety of domains.
Moreover, to gain better insights into the generalization abilities of pre-trained symbolic regression algorithms, the benchmark should include formulas that have not been used in pre-training. 
Since LLMs have most likely seen formulas in publicly accessible web repositories such as Wikipedia, these formulas are not suitable for evaluating the generalization capabilities of pre-trained models.

\subsection{Hybrid Approaches}

Increasing the computational effort of pre-trained symbolic regression algorithms at training time, for instance by using Monte Carlo tree search~ \citep{Shojaee2023TransformerbasedPF,Kamienny2023DeepGS}, has shown potential to find more accurate solutions.
Another idea to combine the two paradigms of search-based and pre-training based symbolic regression is to use a pre-trained algorithm for generating candidate formulas that serve as starting points for search-based symbolic regression algorithms.
The unified framework for deep symbolic regression (uDSR) by~\citet{dsr3_petersen22} uses a transformer, and EvoNUDGE~\citep{evonudge_wyrwinsky24} uses graph neural networks for candidate generation.

\subsection{Equation Discovery Systems}
To address the tendency of symbolic regression algorithms to produce formulas that fit the data well but fail to generalize, a new class of systems has emerged that integrates data-driven discovery with theoretical reasoning to ensure the integrity of discovered laws.

\noindent AI-Descartes~\citep{cornelio2023combining} pioneers a two-stage "generate-then-test" approach. It first uses symbolic regression to generate candidate formulas from data, then employs an automated theorem prover to assess their derivability from a pre-defined background knowledge base. Its key innovation is a "distance to derivability" metric, which quantifies the theoretical consistency of non-derivable formulas to balance empirical fit with theoretical soundness. In contrast, AI-Hilbert~\citep{cory2024evolving} integrates this into a single, unified optimization problem. It simultaneously minimizes both numerical error on data and a distance measure that quantifies derivability from a background knowledge base. By representing laws as polynomials, it leverages the Positivstellensatz~\citep{stengle1974nullstellensatz} theorem to find not only an optimal formula but also a formal, machine-verifiable certificate of its derivation. AI-Newton~\citep{fang2025ai} represents a more foundational paradigm, requiring no pre-supplied background knowledge base. Starting from raw observations, its autonomous workflow iteratively analyzes experiments to build a knowledge base from scratch. Its primary innovation is autonomous concept formation—defining new physical concepts like mass and energy by identifying invariants in data. Through a reasoning component, it then generalizes specific findings into a hierarchical theory, emulating a bottom-up scientific discovery process.\\

\noindent \Forward\, The increasing integration of symbolic regression into larger equation discovery systems necessitates the development of comprehensive diagnostic tools. These tools are vital for identifying the capabilities and limitations of components. 

\section{Equation Recovery Benchmark}
\label{sec:equation_recovery_benchmark}
The \textit{Equation Recovery Benchmark} (ERBench) benchmark aims to evaluate the ability of symbolic regression algorithms to recover ground truth formulas from samples. 
First, we define the equation recovery task. Then, we outline the design principles of the benchmark. After that, we describe the formulas that compose the benchmark and how data is sampled from those formulas. Finally, we discuss the equation recovery challenge and explain how we evaluate the performance of algorithms.

\subsection{The Equation Recovery Task}
\label{sec:equation_recovery_task}
The equation recovery task is motivated by a fundamental challenge in automated scientific discovery: how can we trust an algorithm to discover novel laws we do not yet know? In this work, we address the algorithm selection problem in equation discovery by adopting the principle that the most credible method for discovering new scientific laws is the one that most reliably recovers known ground-truth equations. Consequently, an algorithm’s performance on recovering established formulas across a comprehensive benchmark serves as strong empirical evidence of its potential to uncover new, valid laws of nature. Our benchmark is specifically designed to provide such evidence.

\paragraph{Formal Definition.}
Let $\mathcal{D} = \{(\mathbf{x}_i, y_i)\}_{i=1}^N$ be a dataset of $N$ observations, where $\mathbf{x}_i \in \mathbb{R}^d$ is a vector of $d$ independent variables and $y_i \in \mathbb{R}$ is the corresponding dependent variable. We assume the data is generated from a latent ground truth function $f: \mathbb{R}^d \to \mathbb{R}$, subject to additive noise, such that $y_i = f(\mathbf{x}_i) + \epsilon_i$. The function $f$ is presumed to have a parsimonious symbolic representation.\\

\noindent Classical (symbolic) regression algorithms seek a solution to the regularized empirical risk minimization problem:
\begin{equation}
    \hat{f} = \arg\min_{h \in \mathcal{H}} \left( \frac{1}{N}\sum_{i=1}^N \mathcal{L}(h(\mathbf{x}_i), y_i) + \lambda \Omega(h) \right)
\end{equation}
where $\mathcal{L}$ is a loss function measuring the goodness-of-fit (e.g., mean squared error), and $\Omega(h)$ is a regularization term that penalizes the complexity of the expression $h$ from the hypothesis space $\mathcal{H}$ (e.g., by the number of terms or operators) to enforce parsimony. The hyperparameter $\lambda$ controls the trade-off between data fidelity and model simplicity. The objective of the equation recovery task extends beyond minimizing this objective function.\\

\paragraph{Recovery.} 
The ultimate objective of the equation recovery task is to identify a symbolic expression $\hat{f}$ from a vast hypothesis space that recovers the original symbolic expression $f$, or an expression that is symbolically equivalent to $f$. To ensure robustness against trivial scaling or offset transformations, we adopt a definition of scale or shift equivalence~\citep{srbench_lacava21} and define the recovery goal as success up to scale or shift equivalence, denoted by the relation $\equiv$. A recovered expression $\hat{f}$ is considered equivalent to the ground truth $f$ if it differs only by a constant offset or a non-zero constant scaling factor. Formally, we state $\hat{f} \equiv f$ if and only if a computer algebra system can verify that either:
\begin{equation}
f(\mathbf{x}) - \hat{f}(\mathbf{x})  = c_0 
\quad \text{or} \quad 
\frac{f(\mathbf{x})}{\hat{f}(\mathbf{x})} = c_1
\end{equation}
\noindent for some constants $c_0 \in \mathbb{R}$ or $c_1 \in \mathbb{R} \setminus \{0\}$.
In practice, we use the symbolic mathematics library \texttt{Sympy}~\citep{sympy_meurer17} to perform these simplifications. The overall \textbf{Symbolic Recovery Rate} is the fraction of test problems for which either of these conditions is met.

\noindent We acknowledge that verifying symbolic equivalence is computationally undecidable in the general case~\citep{richardson68}. Consequently, reliance on a Computer Algebra System (CAS) like \texttt{Sympy} implies the risk of false negatives—instances where an algorithm finds a correct but complex representation that the CAS fails to simplify to zero within a given timeout. To mitigate this, we apply a tiered sequence of simplification strategies, including expansion, trigonometric simplification, and rationalization. However, we maintain a strict standard: if the CAS cannot prove equivalence, the attempt is marked as a failure. Therefore, the reported symbolic recovery rates represent a rigorous \textbf{lower bound} on algorithmic performance. To establish an \textbf{upper bound}, we additionally employ a numerical equivalence check.\\  

\noindent We define a predicted expression $\hat{f}$ as numerically equivalent to the ground truth $f$ if, for a set of $k = 1\,000$ randomly sampled validation points $\{x_j\}_{j=1}^k$, the relationship between $\hat{f}$ and $f$ is numerically constant. Specifically, we verify whether the standard deviation of the differences $(f(x_j) - \hat{f}(x_j))$ is within a floating-point tolerance $\epsilon$, or whether the standard deviation of the ratios $(f(x_j) / \hat{f}(x_j))$ is within $\epsilon$ (provided the mean ratio is non-zero). These points are chosen independently of the sample point domains to lie in the interval $[-100, 100]^d$. This metric captures functionally correct solutions that may elude symbolic verification, providing a ceiling for the true recovery rate.\\ 

\noindent In practice, we analyzed the behavior of these bounds and found no difference in the actual values. Hence, we use only the numeric equivalence metric, as it is faster by several orders of magnitude. More details on the analysis can be found in Appendix~\ref{subsec:recovery}.

\paragraph{Relaxed Evaluation Metrics.} 
Recovery is a strict measure of success. Symbolic regression algorithms can fail to reach this goal, but still provide meaningful information, for instance in the form of correctly recovered subexpressions. The \textbf{Jaccard Index (JI)}~\citep{jaccard1901etude} measures the overlap between the set of all unique subexpressions (nodes) of the ground-truth expression tree and the predicted expression tree. Let $T_{\text{true}}$ be the ground-truth tree and $T_{\text{pred}}$ be the predicted tree. Let $S(T)$ be the function that returns the set of all unique nodes from a tree $T$. The Jaccard Index is then computed as follows:
\begin{equation}
    \text{JI}(T_{\text{true}}, T_{\text{pred}}) = \frac{|S(T_{\text{true}}) \cap S(T_{\text{pred}})|}{|S(T_{\text{true}}) \cup S(T_{\text{pred}})|}
    \label{eq:jaccard}
\end{equation}
The JI score ranges from $0$, indicating no common subexpressions, to $1$, indicating that the trees are composed of the exact same set of subexpressions and are therefore symbolically equivalent. For example, if the ground truth is $(x+y) \times z$ and the prediction is $(x+y) + z$, the sets of subexpressions would be $S(T_{\text{true}}) = \{x, y, z, x+y, (x+y)\times z\}$ and $S(T_{\text{pred}}) = \{x, y, z, x+y, (x+y)+z\}$. The JI would be $4/6\approx0.66$, rewarding the algorithm for correctly identifying the $(x+y)$ component.\\

\noindent We also report the \textbf{Tree Edit Distance (TED)}, a structural measure of similarity that although straightforward to define mathematically, is less intuitive to interpret than the Jaccard Index. The TED measures the similarity between the ground-truth expression tree and the predicted expression tree based on the minimum number of edits (inserts, deletes, updates) required to transform one into the other. Both trees are canonicalized before the score is computed. Following the algorithm of~\cite{zhang1989simple} and assuming a unit cost for each edit operation, we define a normalized similarity score as follows:
\begin{equation}
    \text{TED}(T_{\text{true}}, T_{\text{pred}}) = \frac{\text{dist}(T_{\text{true}}, T_{\text{pred}})}{|T_{\text{true}}| + |T_{\text{pred}}|}
    \label{eq:ted}
\end{equation}
Here, $\text{dist}(T_{\text{true}}, T_{\text{pred}})$ is the raw minimum edit distance returned by the Zhang-Shasha algorithm, and $|T|$ denotes the number of nodes in a tree. This score is normalized to the range $[0, 1]$, where 0 signifies that the trees are identical (zero edit distance), and higher values indicate less similarity.
For example, transforming $(x+y) \times z$ to $(x+y) + z$ requires one "update" operation (changing $\times$ to $+$). With 5 nodes in each tree, the TED score would be $\frac{1}{5+5} = 0.1$, reflecting their low edit distance.

\subsection{Design Principles}
\label{sec:design_principles}
As argued in the state-of-the-art review (\Forward), an appropriate benchmark for equation discovery algorithms should cover the following aspects: 

\begin{itemize}
    \item \textbf{Diversity:} The benchmark must be composed of equations from a wide range of scientific domains and mathematical families. This ensures that it does not favor algorithms biased towards a single domain or class of functions. The ERBench suite includes equations from engineering, biology, chemistry, mathematics, physics and synthetic formulas. Further, it covers algebraic (polynomial and rational), and transcendental (e.g., trigonometric, exponential, logarithmic) forms. This allows for a comprehensive assessment of an algorithm's ability to explore a diverse hypothesis space. 
    
    \item \textbf{Generalization:} The focus of the benchmark is the evaluation of performance with respect to ground truth equation recovery. As a hard measure, we employ symbolic equivalency as well two further symbolic soft generalization measures. This enables the quantification and comparison of the generalization abilities of different equation discovery algorithms.
    
    \item \textbf{Robustness:} The benchmark is designed to systematically probe the performance limits of algorithms. It includes problems that scale along several axes of complexity:
    \begin{itemize}
        \item \textit{Dimensionality:} The number of independent variables.
        \item \textit{Complexity:} The number of operators in the true expression.
        \item \textit{Sample Size:} The number of samples in the training data.
        \item \textit{Domain:} The sampling domain of the training and test data. 
        \item \textit{Noise:} The amount of noise applied to the training data. 
        \item \textit{Distribution:} The sampling distribution used to sample the training data.
    \end{itemize}
\end{itemize}

\noindent By adhering to these design principles, the benchmark provides researchers with a comprehensive tool for diagnosing the capabilities and limitations of their algorithms. The benchmark's structure facilitates a granular quantitative analysis, making it possible to assess performance across key axes. Furthermore, it enables a qualitative analysis of an algorithm's ability to recover equations from diverse scientific domains and mathematical function classes. Ultimately, these fine-grained diagnostics provide a robust measure of an algorithm's generalization ability—the most critical indicator of its potential for genuine scientific discovery.

\subsection{Benchmark Components}
\label{sec:benchmark_components}
In order to adhere to core principles of a common task framework~\citep{Liberman10}, we split the benchmark into two datasets: A publicly available development dataset containing ground truth formulas together with samples, and a secret evaluation dataset from which only samples are publicly available.

\paragraph{Public Development Set.}
The public development set enables developers to optimize hyperparameters of their equation discovery algorithms. It contains a total of $10\,000$ ground truth formulas together with sampling domains on which the formulas can be evaluated. 
The formulas, listed in Table~\ref{tab:datasets}, come from various sources: 
We started with $606$ legacy formulas that have been used before to assess the performance of symbolic regression algorithms. 
This collection includes the Feynman and Strogatz equations~\citep{feynmanAI_udrescu20, strogatz_problems}, which are used by SRBench.
However, since these formulas only cover a limited range of domains, we collected $4001$ additional formulas from the natural sciences, collections of probability density functions, Wikipedia, and collections of integer sequences.
Finally, we added $5\,303$ synthetic formulas that were generated by sampling expression DAGs. Similar methods for synthetic equation generation have been proposed in~\citep{udfs_kahlmeyer24} and~\citep{lample2019deep}.
We call this dataset \textbf{Syn}thetic \textbf{eq}uation set (SynEq). A detailed description of the sampling process including hyperparameters is provided in Appendix~\ref{subsec:creating_syneq}.\\ 

\noindent \textbf{Access to the public development set}: \includegraphics[width=10pt]{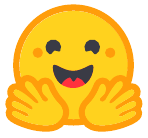} \,\href{https://huggingface.co/datasets/EquationDiscovery/Equation_Recovery_Benchmark}{Huggingface}

\begin{table}[t!]
\caption{\textbf{Composition of the public development set of ERBench.} The development set comprises datasets from various fields: engineering (E), biology (B), chemistry (C), mathematics (M), physics (P), and synthetic data sets (S). Datasets marked with an asterisk are new and have been collected or compiled by the authors. For all datasets we provide the license of the source.\\}
    \centering
    \begin{tabular}{lllll}
         Name&Instances&Source&Fields&License\\
         \toprule
         \textbf{Densities*}&33&SciPy~\citep{scipy}&EMP&BSD 3-Clause\\
         \textbf{Eponymous*}&211&\citet{eponymous}&EBCMP&CC-BY-SA 4.0\\
         Feynman&130&\citet{feynmanAI_udrescu20}&P&MIT\\
         Keijzer&15&\citet{keijzer2003}&S&BSD 3-Clause\\
         Korns&15&\citet{korns2011}&S&BSD 3-Clause\\
         Koza&2&\citet{koza1994genetic}&S&BSD 3-Clause\\
         Livermore&175&\citet{dsr2_petersen21}&S&BSD 3-Clause\\
         Nguyen&12&\citet{nguyen2011}&S&BSD 3-Clause\\
         \textbf{OEIS*}&3\,757&\citet{oeis}&M&CC BY-SA 4.0\\
         Pagie&1&\citet{pagie1997}&S&BSD 3-Clause\\
         \textbf{PHYBench*}&90&\citet{qiu2025phybench}&P&MIT\\
         \textbf{SynEq*}&5\,303&this paper &S&MIT\\
         SRDS&234&\citet{matsubara2024srds}&P&MIT\\
         Strogatz&14&\citet{strogatz_problems}&BP&GPL-3.0\\
         Vladislavleva&8&\citet{vladislavleva2009}&S&BSD 3-Clause\\
         \midrule
         \textbf{Total}&10\,000&&&
    \end{tabular}
    \label{tab:datasets}
\end{table}

\paragraph{Secret Evaluation Set.}
The evaluation set is used to assess an equation discovery algorithm's ability to discover ground truth formulas. 
We use a deterministic, parameterizable formula generation process that generates a set of $1\,000$ formulas. The formula generation method as well as the held out set for the competition are deliberately kept secret.\\

\noindent \textbf{Access to competition}: \includegraphics[width=10pt]{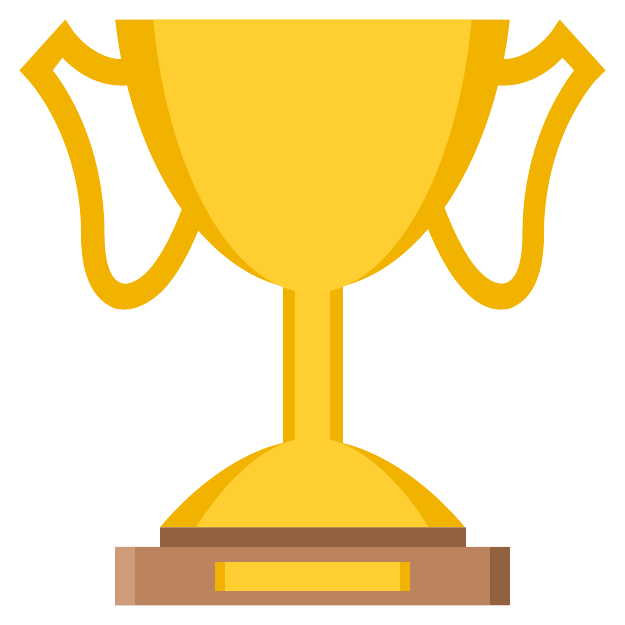} \,\href{https://equation-discovery.ti2.fmi.uni-jena.de/}{Competition}

\paragraph{Data Generation Protocol.}
For each formula from the development and from the evaluation set we generate samples, that is, argument-label pairs $(x,y)$, that can be used for training the equation discovery algorithms.
To do so, we need to define sampling domains for the formulas from the benchmark data sets.
The legacy formulas and the formulas from the \emph{Densities} data set come with their own sampling domains, and we keep these domains.
The formulas in the \emph{Eponymous} dataset are scraped from the Wikipedia article with the same name using the \texttt{mistral-large-2411} LLM~\citep{mistral7b}.
The LLM also provides us with suitable sampling domains, which have been cross-checked by the authors. The same procedure was applied to the \emph{PHYBench} problems.
For the formulas in the \emph{OEIS} dataset, we always used the first 500 integers. 
For the \emph{SynEq} dataset, we determine a large box in $\mathbb{R}^n$, where $n$ is the number of features, on which the formulas are valid, that is, we avoid division by zero, and taking logarithms or square-roots of negative numbers. For each formula, to draw samples from the sampling domains, we use the diverse sampling distribution that has been introduced by~\citet{kamienny22_transformer}. 
We provide a more detailed description of the sampling strategy in Appendix~\ref{subsec:creating_syneq}.

\subsection{Equation Recovery Challenge}
\label{sec:equation_recovery_challenge}
\begin{figure}[t!]
     \centering
     \begin{minipage}{0.495\textwidth}
        \centering
        \input{figures/eval_protocol1.tex}
     \end{minipage}
     \hfill
     \begin{minipage}{0.495\textwidth}
        \centering
        \input{figures/eval_protocol2.tex}
     \end{minipage}\\[4pt]
    \caption{\textbf{Communication protocol for the secret evaluation set.} 
    \textbf{Left:} On user request, the website generates a unique request ID that serves as a seed for permuting the evaluation set. The user then has access to the permuted regression problems (without ground truth) and the submission ID. 
    \textbf{Right:} Once the user submits expressions and the request ID, the website re-generates the permuted evaluation set and calculates the evaluation metrics. Only aggregated scores are returned to the user.}
    \label{fig:eval_protocol}  
\end{figure}
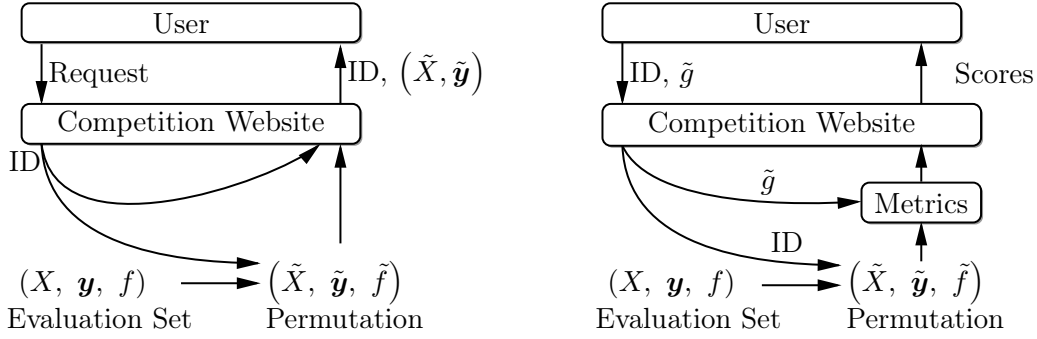

Formulas from  the public development set can be used to pre-train equation discovery algorithms and for evaluating algorithms privately, that is, locally by the users. 
For public evaluation and benchmarking, we cannot use the public development data set, because of potential overfitting to this data set.
Therefore, we need the secret evaluation set of $1\,000$ formulas that are accessible to users only through samples.
Users train their equation discovery algorithms on these samples and submit the resulting formulas for evaluation to our competition website. 
In the following, we describe the workflow of the competition website. 
The workflow, which is illustrated in Figure~\ref{fig:eval_protocol}, has three phases: Requesting problems, running the equation discovery algorithm, and submitting and evaluating the results.

\paragraph{Requesting Problems.}
To maintain the integrity of the secret test set and prevent users from revealing the ground truth via repeated queries, we implement a strict permutation strategy. For each user request, a new sample is generated from the formulas of the evaluation data set. However, simply re-sampling is insufficient; we must ensure the samples provide equally difficult regression problems while remaining uninformative across multiple requests. To achieve this, we permute the data on three levels: (1) \textbf{Problem order:} The sequence of regression tasks is shuffled.
(2) \textbf{Sample order:} Within each problem, the row-wise order of sample points $(x_i, y_i)$ is shuffled.
(3) \textbf{Variable order:} The columns of the independent variables (features) are permuted.\\

\noindent For a simple example, consider an evaluation set with two problems with two feature variables each, 
a problem with three samples of the formula $x_1\cdot x_2$ and a problem with two samples of the formula $x_2-x_1$. 
The problems are given as tuples of triples $(X, \boldsymbol{y}, f)$, with $f(x_i) = y_i$, before (left) and after (right) permutation:
\begin{align*}
{\tiny
    \Bigg(\begin{bmatrix}
    1&2\\
    3&4\\
    5&6
    \end{bmatrix},\begin{bmatrix}
        2\\
        12\\
        30
    \end{bmatrix}, x_1x_2\Bigg), 
    \Bigg(\begin{bmatrix}
    1&2\\
    2&4\\
    \end{bmatrix},\begin{bmatrix}
        1\\
        2\\
    \end{bmatrix}, x_2-x_1\Bigg)\rightarrow
    \Bigg(\begin{bmatrix}
    4&2\\
    2&1
    \end{bmatrix},\begin{bmatrix}
        2\\
        1
    \end{bmatrix}, x_1-x_2\Bigg),
    \Bigg(\begin{bmatrix}
    1&2\\
    5&6\\
    3&4\\
    \end{bmatrix},\begin{bmatrix}
        2\\
        30\\
        12
    \end{bmatrix}, x_1x_2\Bigg)
}
\end{align*}

\noindent On user request, the website generates a unique request ID that is used to seed a pseudo-random number generator that is then used to compute request-specific random permutations.
For the generated request ID, the permuted problems are still triples $(\tilde{X}, \tilde{\boldsymbol{y}}, \tilde{f})$ from which the samples $(\tilde{X}, \tilde{\boldsymbol{y}})$ are passed, together with the request ID, to the user.
The procedure is illustrated in Figure~\ref{fig:eval_protocol} (left).

\paragraph{Submitting Results.}
The samples $(\tilde{X}, \tilde{\boldsymbol{y}})$ of the permuted problem provide training data for the user's equation discovery algorithm. 
On input of these training data, users' algorithms compute formulas $\tilde{g}$ that should recover the formulas $\tilde{f}$ from the permuted evaluation set.
The formulas $\tilde{g}$ are then sent, together with the request ID, to the competition website for evaluation.
For the evaluation on the competition website, the request ID is used to compute the request-specifically permuted formulas $\tilde{f}$ that can be compared for symbolic equivalence to the formulas $\tilde{g}$.
The procedure is illustrated in Figure~\ref{fig:eval_protocol} (right).\\

\noindent In contrast to other benchmarks like, e.g. SRBench, the equation discovery algorithms for ERBench are run by the users themselves.
This has three advantages: First, competing equation discovery algorithms no longer have to adhere to the same interface in a predefined programming language, second, the competition runs continuously and not only at fixed, compute-heavy events, and third, the competition website is lightweight and easily maintainable. 

\paragraph{Evaluation.}
Once the request-specific permuted formulas $\tilde{f}$ have been recomputed from the formulas $f$ in the evaluation set and the request ID, the aforementioned performance metrics are computed on the competition website. 
In addition to performance metrics, we record the \emph{time-to-submission}, defined as the duration between the dataset request and the result submission. This allows for an indirect comparison of the computational time required by competing algorithms to produce benchmark results, providing insight into their time complexity. It is, however, only a proxy, because the competitors can use different hardware setups.
In order to prevent overfitting to the requested problems, the website only returns performance scores that have been aggregated over the evaluation set. 
Moreover, the request ID is invalidated by the website, that is, the samples associated with a request ID can only be used for evaluation once. 

\section{Experiments}
To quantitatively assess the current state of the art in equation discovery, we conducted a rigorous experimental evaluation. This section details our methodology and findings. We begin by describing the experimental setup, including the selection of representative algorithms and the evaluation protocol. We then present the aggregate performance results on the private test set, providing the first direct comparison of these methods on a standardized task. Finally, we use the benchmark's diagnostic tools for a more fine-grained analysis of the top-performing algorithm, illustrating how ERBench helps developers investigate their specific algorithm's strengths and weaknesses.

\subsection{Experimental Setup}
\label{sec:experimental_setup}
In the following, we introduce the evaluated algorithms and the baselines used for comparison, as well as the experimental setup employed in our evaluation.

\paragraph{Algorithms.}
We benchmark a diverse set of prominent equation discovery algorithms, selected to represent the major paradigms in the field:
\begin{itemize}
    \item \textbf{Search-based algorithms:} We include \texttt{PySR} \citep{pysr_cranmer23}, a state-of-the-art framework for symbolic regression using multi-population evolution, alongside \texttt{Operon} \citep{operon_20} and \texttt{gplearn} \citep{gplearn_stephens16}, which are both strong and widely recognized genetic programming implementations.
    \item \textbf{Pre-trained algorithms:} We evaluate \texttt{E2E} \citep{kamienny22_transformer}, a transformer-based model designed for rapid inference after a one-time pre-training phase.
    \item \textbf{Reinforcement learning algorithms:} We report the performance of \texttt{DSR}~\citep{dsr_petersen21}, a representative reinforcement learning–based method that learns symbolic expressions via online policy-gradient optimization on each target dataset.
\end{itemize}
For all methods, we utilized their official, publicly available implementations. To ensure a fair and standardized comparison, we employed the default hyperparameter configurations provided by the authors in their respective repositories. This approach evaluates the out-of-the-box performance of each algorithm, eliminating variability introduced by extensive hyperparameter tuning.

\paragraph{Baseline.}
To establish a performance floor, we include standard linear regression as a baseline model. This baseline helps to contextualize the performance of the more complex equation discovery algorithms.

\paragraph{Evaluation Protocol.}
All experiments were conducted on a server equipped with an Intel Xeon Gold 6226R CPU. The complete code base for our evaluation and for reproducing all results is publicly available.\footnote{URL: \url{https://huggingface.co/datasets/EquationDiscovery/Equation_Recovery_Benchmark}} 
For each problem within the ERBench evaluation set, every algorithm was tasked with recovering the underlying equation from the provided data. Upon completion, the final symbolic expression predicted by each algorithm was evaluated against the ground-truth equation. As defined in Section~\ref{sec:equation_recovery_task}, our primary metrics are the \emph{Symbolic Recovery Rate}, \emph{Jaccard Index (JI)}, and the \emph{Tree Edit Distance (TED)}. Secondary metrics, such as predictive accuracy on a hold-out test set, were also computed. The aggregated statistics over the entire benchmark were then used to compare the overall performance of the algorithms.

\subsection{Results}
\label{sec:results}
In the following, we compare the performance of the individual algorithms and discuss the insights obtained from our diagnostic analysis. 

\begin{table}[]
    \small
    \centering
        \caption{
        \textbf{Aggregate performance of state-of-the-art symbolic regression algorithms on ERBench.} 
        Values represent the mean $\pm$ standard deviation over five independent runs. 
        \textbf{Recovery} is the fraction of problems where the ground-truth expression was recovered exactly (higher is better). 
        \textbf{Jaccard Index} measures the similarity of symbolic subsets between the predicted and true equations (higher is better). 
        Tree Edit Distance (\textbf{TED}) measures the number of edit operations required to transform the predicted expression tree into the ground truth (lower is better). 
        The results show a significant performance gap, with PySR being the only method to achieve a non-trivial symbolic recovery rate. 
    }
    \begin{tabular}{lllllll}
    &PySR&DSR&E2E&Operon&gplearn&Linear\\
    \toprule
    Recovery&$0.29\pm0.04$&$0.0\pm0.0$&$0.0\pm0.0$&$0.0\pm0.0$&$0.05\pm0.01$&$0.0\pm0.0$\\
    Jaccard Index&$0.49\pm0.02$&$0.28\pm0.01$&$0.17\pm0.0$&$0.13\pm0.0$&$0.27\pm0.0$&$0.25\pm0.0$\\
    TED&$13.4\pm0.7$&$17.6\pm0.4$&$30.4\pm0.2$&$37.0\pm0.6$&$45.2\pm2.1$&$14.6\pm0.0$\\
    \end{tabular}

    \label{tab:erbench_results_table}
\end{table}

\paragraph{Benchmark Performance.}
The aggregate results of our evaluation, presented in Table~\ref{tab:erbench_results_table}, underscore the  challenge posed by ERBench to current state-of-the-art algorithms. 
Focusing on the strict metric of symbolic recovery, the benchmark proves exceptionally difficult for most methods. Five of the six algorithms, including the transformer-based \texttt{E2E} and established genetic programming methods like \texttt{Operon}, achieve a recovery rate near zero. In stark contrast, \texttt{PySR} successfully recovers $\sim30\%$ of the equations, establishing a clear performance, confirming its state-of-the-art status in exact symbolic discovery.\\

\noindent The Jaccard Index, which measures partial symbolic similarity, provides a more nuanced view. \texttt{PySR} again leads with a dominant score of $0.5$, indicating that on average, its incorrect solutions still share half of their symbolic components with the ground truth. More surprisingly, the performance of the next best algorithm, \texttt{DSR} ($0.28$), is closely matched by the simple Linear baseline ($0.25$). This notable result suggests that complex algorithms often generate large expression trees with many incorrect sub-expressions. Since the Jaccard Index penalizes the union of non-overlapping sub-expressions (the denominator in Eq.~\ref{eq:jaccard}), the compact nature of the \texttt{Linear} baseline (which introduces no extraneous operators) yields a higher score than methods that produce complex, incorrect mathematical structures. This highlights that current state-of-the-art methods struggle not only with finding the truth but with maintaining parsimony as well.

\paragraph{Diagnostic Analysis.}
Beyond the aggregate performance metrics from the private test set, ERBench provides a suite of diagnostic tools designed for fine-grained analysis on its public datasets. This allows researchers to move beyond simple rankings to understand the specific strengths and weaknesses of their methods. To demonstrate the utility of this suite, we conduct a detailed analysis of the top-performing algorithm, \texttt{PySR}, on the Feynman dataset subset of our collection.\\

\begin{figure}[t!]
    \centering 

    \begin{subfigure}[b]{0.49\textwidth}
        \centering
        \includegraphics[width=\linewidth]{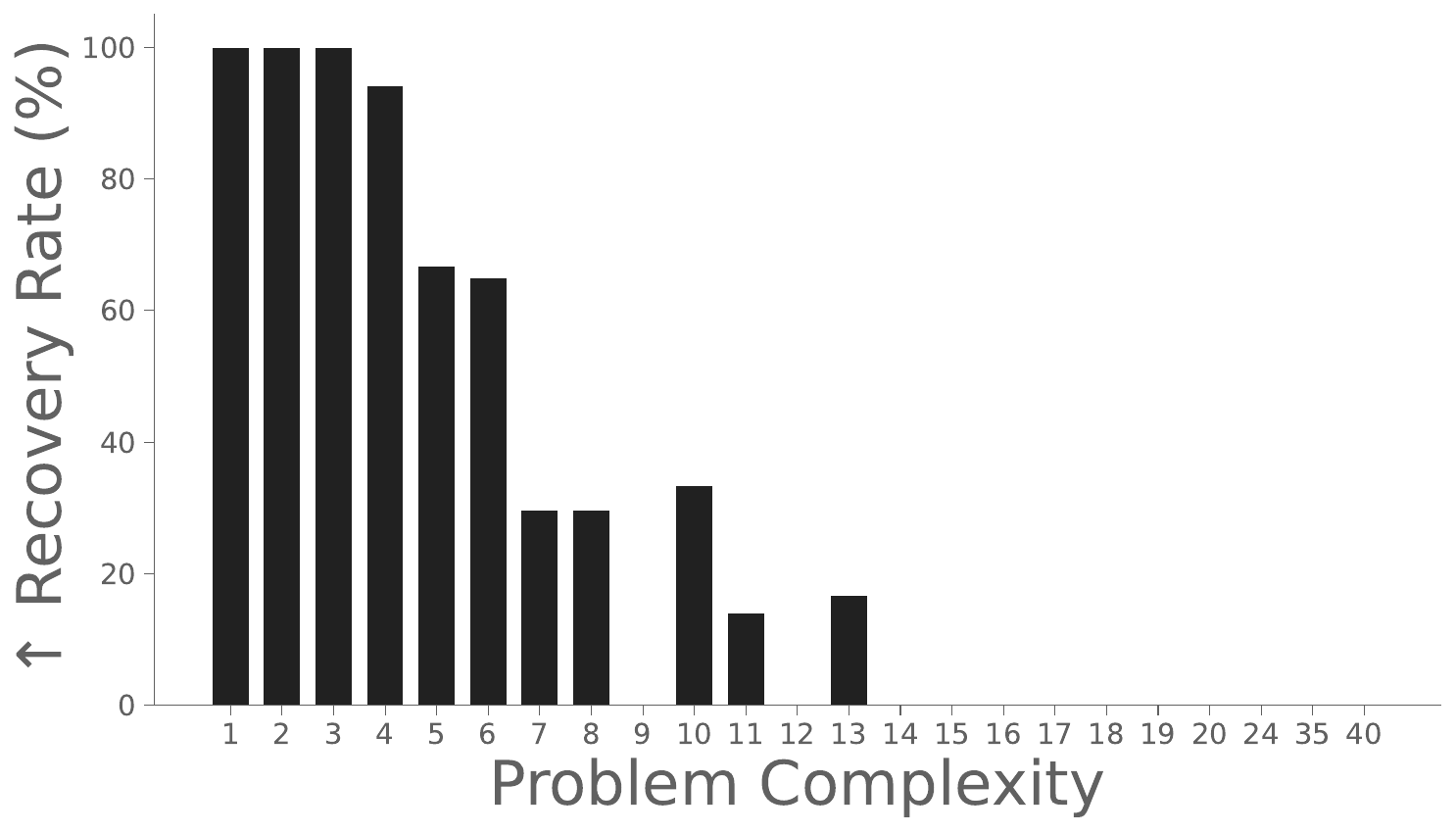}
        \caption{} 
        \label{fig:sub1}
    \end{subfigure}%
    \hfill 
    \begin{subfigure}[b]{0.49\textwidth}
        \centering
        \includegraphics[width=\linewidth]{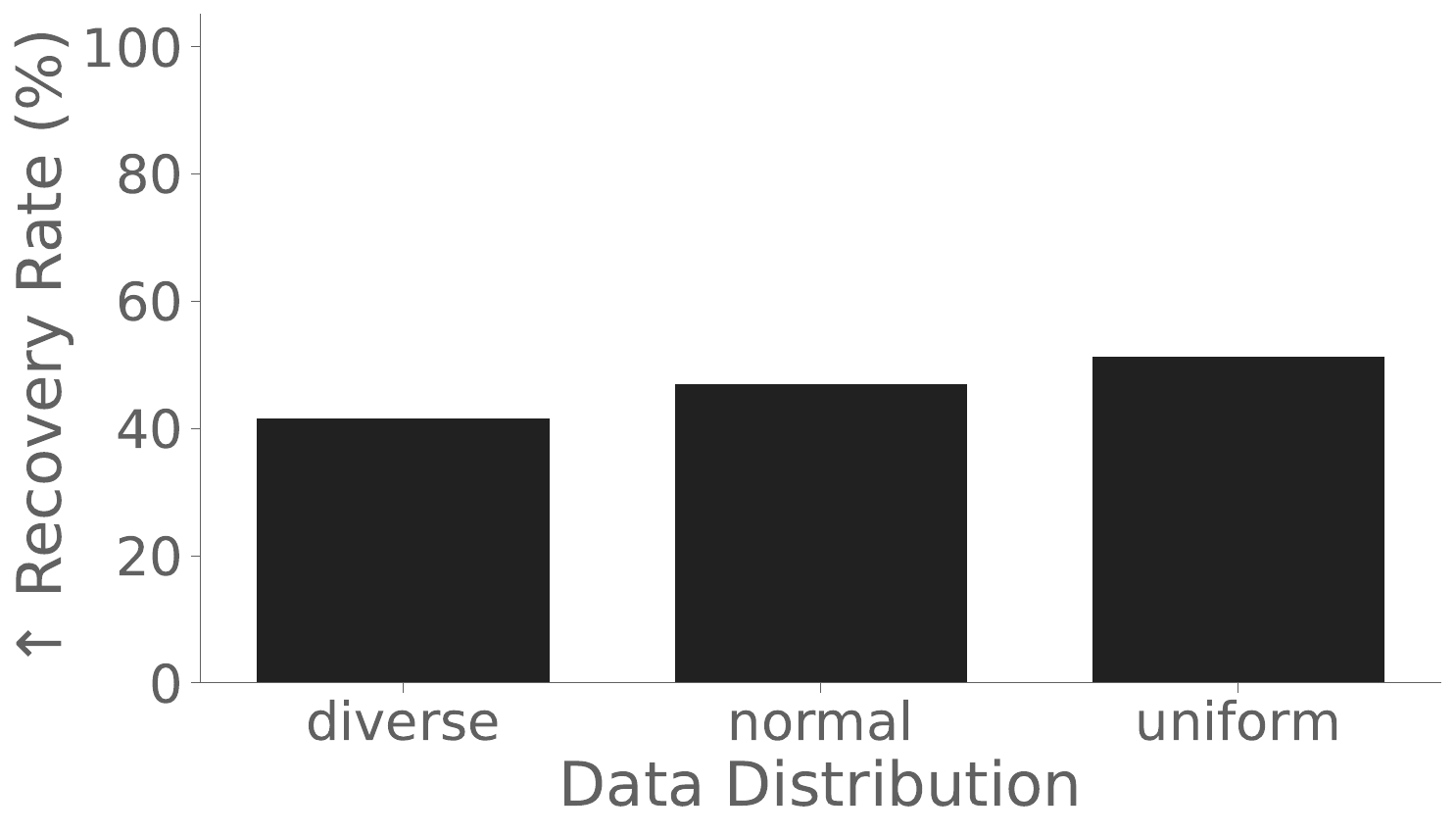}
        \caption{} 
        \label{fig:sub2}
    \end{subfigure}
    
    \vspace{1em} 

    \begin{subfigure}[b]{0.49\textwidth}
        \centering
        \includegraphics[width=\linewidth]{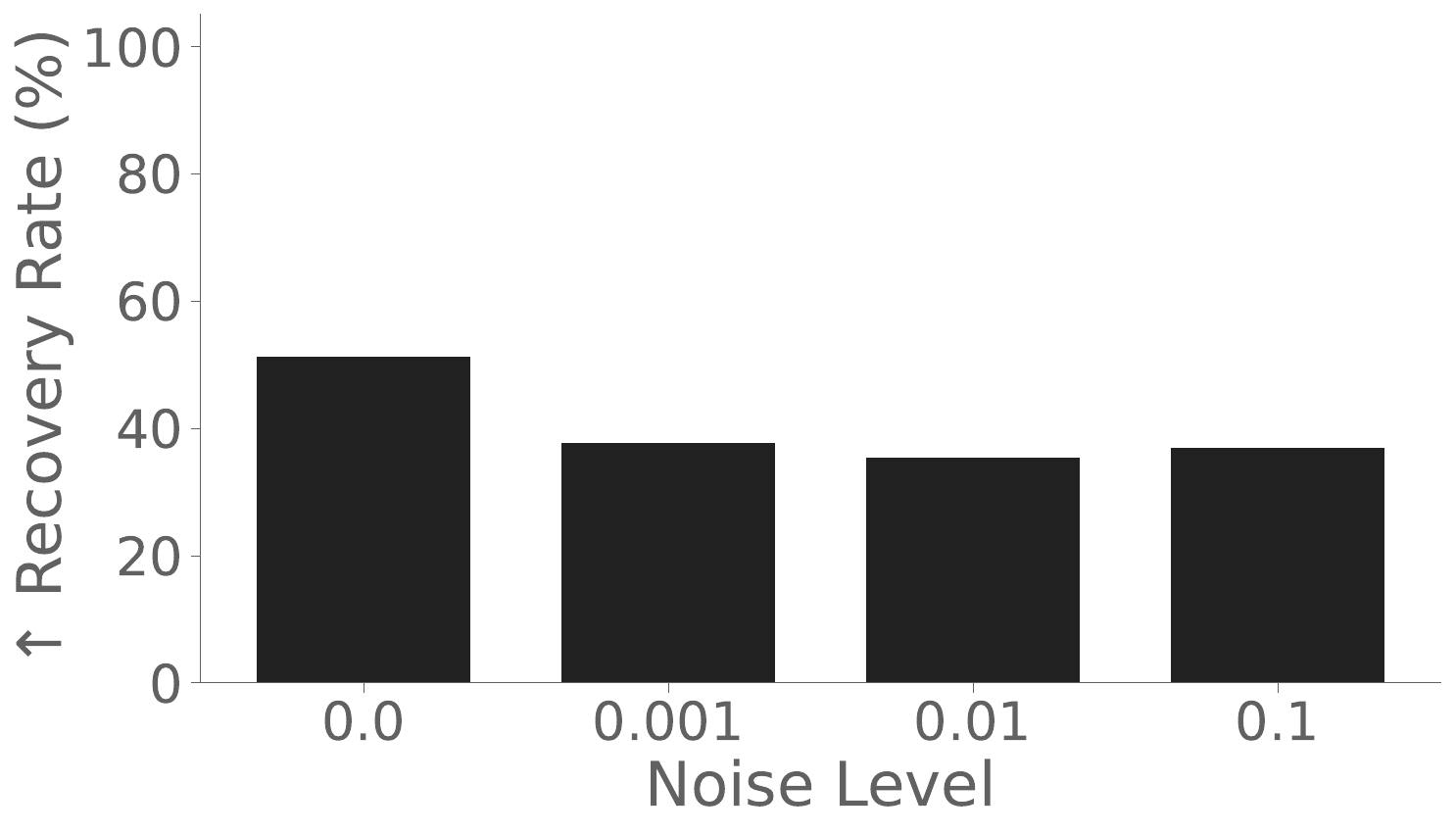}
        \caption{} 
        \label{fig:sub3}
    \end{subfigure}%
    \hfill 
    \begin{subfigure}[b]{0.49\textwidth}
        \centering
        \includegraphics[width=\linewidth]{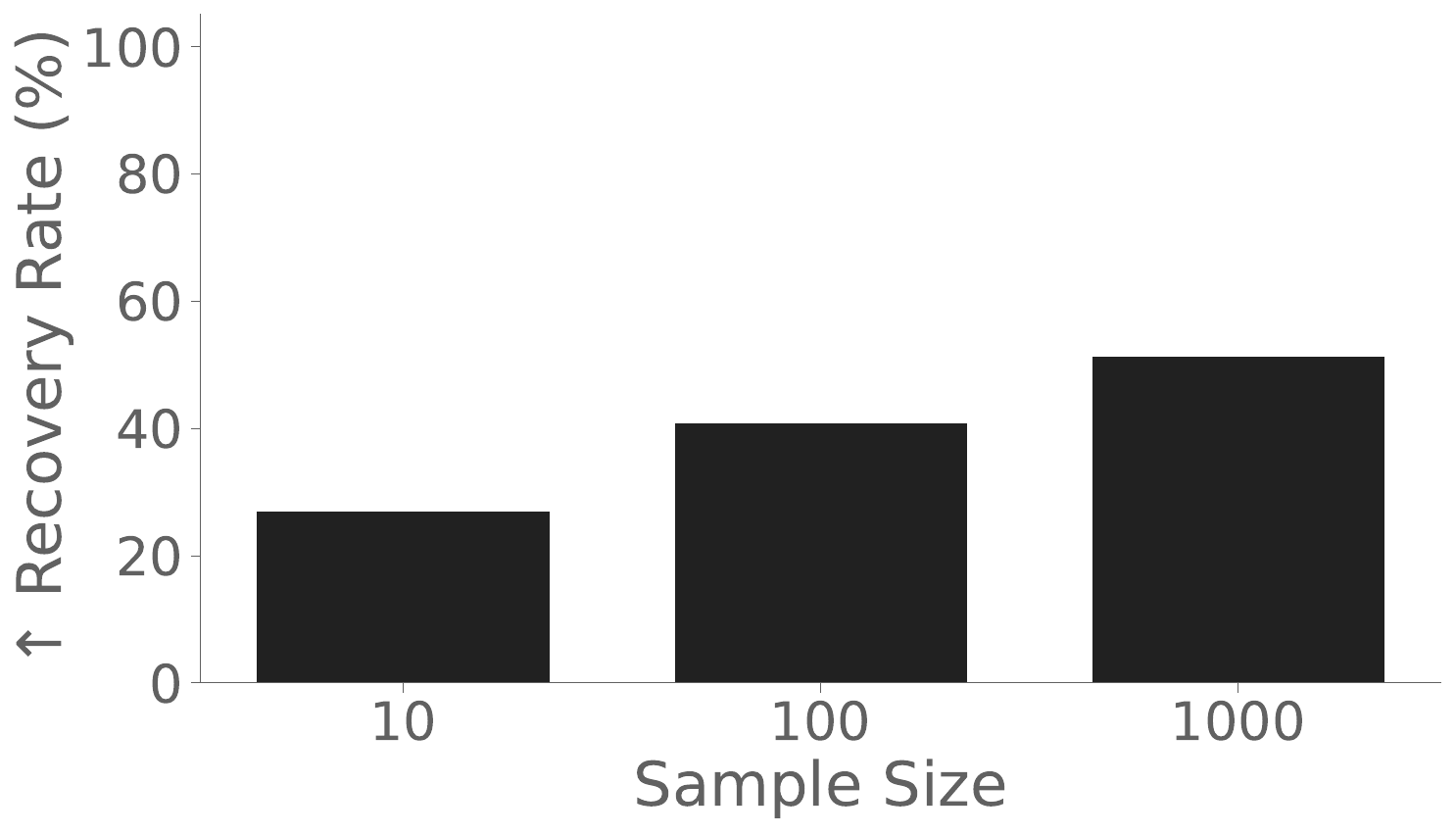}
        \caption{} 
        \label{fig:sub5}
    \end{subfigure}
    
    \caption{
        \textbf{Diagnostic analysis of PySR's performance.} 
        The plots reveal the specific conditions under which the algorithm succeeds or fails.
        \textbf{(a)} Recovery rate against problem complexity (number of operators). 
        \textbf{(b)} Performance trend with respect to data distribution.
        \textbf{(c)} Performance trend with respect to noise level.
        \textbf{(d)} Performance trend with respect to sample size.
        While PySR achieves near-perfect recovery on low-complexity problems, performance degrades as complexity increases. This multi-faceted analysis allows developers to pinpoint the strengths and weaknesses of the algorithm.
    }
    \label{fig:diagnostic_analysis}
\end{figure}

\noindent The results of this analysis are visualized in Figure~\ref{fig:diagnostic_analysis}. 
The first plot, which charts the recovery rate against equation complexity, reveals a stark trend. \texttt{PySR} excels on equations with low complexity ($1$-$3$ operators), achieving near-perfect recovery. However, its performance deteriorates rapidly as the number of operators increases, dropping to near zero for problems with more than $13$ operators. 
The second plot shows that when data is sampled from a uniform distribution, the recovery rate is up to approximately $10$\% higher than when the data follows the diverse distribution introduced by~\citet{kamienny22_transformer}.
The third plot demonstrates that as noise increases, the recovery rate steadily decreases.
In contrast, increasing the sample size leads to a steady improvement in recovery, with performance improving as more samples are provided to the algorithm. This multi-faceted view pinpoints that problem complexity, rather than data scarcity or moderate noise, is the principal bottleneck for this state-of-the-art method, providing a clear target for future research. Such detailed feedback is invaluable for identifying specific limitations and guiding the development of more capable algorithms.

\section{Discussion}
In this section, we discuss the broader implications of our experimental findings. We begin by analyzing the results to identify key challenges and outline promising directions for future research in equation discovery. Following this analysis, we address the limitations of the ERBench benchmark to provide a complete picture of its scope and intended utility for the research community.

\paragraph{Directions for Future Research.}
Our evaluation on ERBench provides the first standardized snapshot of the capabilities and limitations of modern equation discovery algorithms. The results show that the near-zero symbolic recovery rates for most methods in Table~\ref{tab:erbench_results_table} demonstrate that robustly discovering exact equations from data remains a largely unsolved problem. Our diagnostic analysis of the top-performing algorithm, \texttt{PySR}, highlights a primary bottleneck: a rapid decline in performance as problem complexity increases (Figure~\ref{fig:diagnostic_analysis}).

Based on these findings, we identify two critical directions for future research. First, there is a clear need for more elaborate and better informed search strategies to navigate the vast space of symbolic expressions. Such a leap can then allow methods to scale to expressions of higher complexity. Promising ideas could involve developing a better representation for the space of symbolic expressions, or methods that explicitly search for compositional or modular structures. Second, the fact that several complex algorithms were competitive with or even outperformed by a simple linear baseline on the Jaccard Index highlights a fundamental issue of failed guidance. Most algorithms are designed to produce a good fit for given training data points. This makes them good curve-fitters, but poor at recovering equations. They then consequentially act at the same level as a simple linear baseline. Future work should focus on improving the robustness of the search process, perhaps by developing methods that can identify and build upon partially correct sub-expressions, rather than being misled by model fit and then failing entirely on recovery. 

\paragraph{Limitations of ERBench.}
In contrast to SRBench, for evaluating symbolic regression algorithms on ERbench the algorithms are run by the users themselves.
A drawback of this design decision is that the computation times of algorithms from different users are no longer directly comparable. However, we believe that this drawback is outweighed by at least three advantages:  First, competing symbolic regression algorithms no longer have to adhere to the same interface in a predefined programming language, second, the competition is running continuously and not only at fixed, compute-heavy events, and third, the competition website becomes lightweight and easily maintainable.

While ERBench was designed to provide a diverse set of problems from multiple scientific domains, no benchmark can be truly comprehensive. The current version may lack extensive coverage of equations from certain fields, such as finance and economics. We view ERBench as an evolving resource and intend to expand its domain coverage in future iterations, welcoming contributions from the community to enhance its breadth and relevance.

Regarding the private test set, its contents are intentionally withheld to ensure the integrity of future evaluations. Nonetheless, we acknowledge that, like any benchmark, ERBench is subject to inherent selection biases. The distribution of problems, though curated for diversity, may inadvertently favor certain algorithmic paradigms over others. We therefore strongly encourage researchers to use the public dataset and its accompanying diagnostic tools for detailed ablation studies, as this provides a more complete picture of an algorithm's capabilities than any single performance score.

\section{Conclusion}
In this work, we introduced ERBench, a comprehensive benchmark designed to standardize the evaluation and accelerate the development of equation discovery algorithms. We proposed a set of principled primary metrics, namely, Symbolic Recovery Rate, Tree Edit Distance, and Jaccard Index, to move beyond predictive accuracy and directly measure the core task of recovering ground-truth symbolic expressions. ERBench is structured as a common task framework, with a large public dataset for development and a private test set for objective evaluation, ensuring fair and reproducible comparisons. Our initial evaluation of state-of-the-art methods on ERBench provided a baseline: robust symbolic recovery remains a significant challenge, with performance degrading rapidly as problem complexity increases. This highlights the need for this benchmark. By providing a common platform and a suite of diagnostic tools for analyzing performance across dimensions like noise, complexity, and data scarcity, ERBench offers researchers the infrastructure needed to identify algorithmic weaknesses and innovate effectively. We hope that this benchmark will foster the development of strong, robust, and scalable equation discovery algorithms, bringing us closer to the goal of truly automated scientific discovery.

\section*{Broader Impact Statement}
This work introduces ERBench, a benchmark designed to evaluate and advance the field of equation discovery. The primary societal impact of this research is accelerating scientific discovery. ERBench provides a rigorous framework for validating symbolic regression algorithms, enabling the development of AI systems that can uncover interpretable and generalizable laws from data. Unlike "black-box" deep learning models, the symbolic models promoted by this benchmark are inherently interpretable, enabling scientists to inspect, verify, and trust the mechanisms that govern predictions. This directly contributes to the development of trustworthy AI in the natural sciences.

However, we acknowledge the risks and responsibilities associated with this line of research. First, the evaluation of symbolic regression algorithms, particularly those that rely on genetic programming or large pre-trained transformers, can be computationally intensive, which is why environmental impact is a concern. By establishing a standardized benchmark with a fixed evaluation protocol, we aim to reduce redundant training cycles and streamline resource usage in the evaluation process.

Second, regarding bias and scope, ERBench covers diverse domains such as physics, biology, and engineering. However, the ground-truth formulas are limited to known mathematical forms and existing scientific knowledge. There is a risk that algorithms optimized solely on this benchmark will become biased toward discovering clean, parsimonious laws similar to those in the dataset. These algorithms may then struggle with messy, high-dimensional, or non-analytical relationships found in fields such as the social sciences or economics. We mitigate this risk by including a diverse synthetic dataset and encouraging the community to contribute new domains.

Finally, regarding automation bias, there is a risk that practitioners may rely too heavily on automated equation discovery systems and accept the discovered formulas as absolute truth without experimental validation. Our paper emphasizes falsifiability and out-of-domain generalization as core metrics. It explicitly encourages a workflow in which generated equations are treated as hypotheses that are subject to rigorous empirical testing, rather than as final answers.

\section*{Reproducibility Statement}
To ensure the reproducibility of our benchmark and the experimental results presented in this paper, we have made all resources publicly available and thoroughly documented our experimental procedures.

\paragraph{Data Availability.}
The \textit{Public Development Set}, comprising 10,000 ground-truth formulas and their corresponding metadata, is hosted on Hugging Face at \url{https://huggingface.co/datasets/EquationDiscovery/Equation_Recovery_Benchmark}. This repository includes the datasets listed in Table~\ref{tab:datasets}, along with their respective licenses. The \textit{Secret Evaluation Set} is withheld to prevent overfitting but is accessible for evaluation purposes through our competition website at \url{https://equation-discovery.ti2.fmi.uni-jena.de/}.

\paragraph{Code Availability.}
All source code required to replicate the benchmark creation and the experimental evaluation is available in the associated repository on Hugging Face. This includes:
\begin{itemize}
    \item The evaluation scripts implementing the Symbolic Recovery, Jaccard Index, and Tree Edit Distance metrics.
    \item The sampling algorithms for the diverse distribution strategy.
    \item Wrappers for the baseline algorithms (PySR, DSR, E2E, Operon, gplearn) to ensure standardized execution.
\end{itemize}

\paragraph{Experimental Reproduction.}
To reproduce the results in Section~\ref{sec:results} (Table~\ref{tab:erbench_results_table}) and the diagnostic analysis (Figure~\ref{fig:diagnostic_analysis}):
\begin{itemize}
    \item We provide a script for evaluation in the Hugging Face repository. This script includes the necessary baselines and executes the evaluation protocol.
    \item All experiments were conducted using the default hyperparameters provided by the respective algorithm authors, as detailed in Table~\ref{tab:param_settings} (Appendix~\ref{secA1}), ensuring fair out-of-the-box comparison.
\end{itemize}

\paragraph{Maintenance.}
The authors are committed to maintaining the competition website and the Hugging Face repository. As the field progresses, we will facilitate the addition of new datasets and baselines.

\bibliography{bibliography}

\appendix

\newpage 

\section{Additional Details}\label{secA1}

\subsection{Recovery}
\label{subsec:recovery}
In Section~\ref{sec:equation_recovery_task}, we discussed the evaluation metrics that we use to compare the interpretability of different algorithms. In particular, we use symbolic recovery, that is a direct check for symbolic equivalence. For a given ground truth function $f$ and an estimated function $g$, we verify that $f-g$ or $f/g$ simplifies to a constant using \texttt{SymPy}. Since these checks can take quite some time, we also discuss a numeric alternative: On 1\,000 random data points in $[-100, 100]^d$ we evaluate the expressions $f-g$ and $f/g$ and verify that the variance of the function values is close to zero.

The pure symbolic check can produce false negatives (two equivalent functions are labeled as not equivalent), but no false positives (two non-equivalent functions are labeled as equivalent). Contrary, the numeric check can produce false positives but no false negatives. Consequently, the symbolic recovery rate is a lower bound for the true recovery rate. Similarly, the numeric recovery rate is an upper bound for the true recovery rate. 
But how far is that upper bound from the lower bound in practice? 
In order to compare the accuracy of these two approaches, we framed the equivalence test as a classification task. Given a function $f$, we construct two functions $g_0$ and $g_1$ so that $g_0$ can be transformed into $f$, but $g_1$ cannot. We then collect a large dataset of these examples and track how many times the symbolic and numeric checks correctly identify the relation. The equivalent function $g_0$ was created using \texttt{SymPy} (e.g. by applying \texttt{expand, factor, cancel, apart}). In order to create $g_1$, we replaced $x_0$ with $x_0+1$ in the original function.
For the Feynman Problems, both methods achieved 100\% accuracy on the created dataset, which means that the numeric version is very accurate. However, while the symbolic checks took 2.55 seconds on average, the numeric checks took 0.03 seconds on average. Thus, we concluded to use the numeric version in our implementation.

\subsection{State-of-the-Art Experiments}
\label{subsec:sota_exp}
In Section 2 of the paper, we discuss strengths and weaknesses of existing symbolic regression algorithms. Some aspects are supported by small experiments on state-of-the-art symbolic regression algorithms. 
In order to let the experiments run in reasonable time, we used the parameters listed in Table~\ref{tab:param_settings}.
All experiments were run on a computer with an Intel Xeon Gold 6226R 64-core processor, 128 GB of RAM, and Python 3.10.

\begin{table}[ht]
    \small
    \begin{tabular}{lll}
        \toprule
        UDFS~\citep{udfs_kahlmeyer24} 
        & DSR~\citep{dsr_petersen21} 
        & PySR\citep{pysr_cranmer23} \\
        \midrule
        \texttt{n\_consts=1}          & \texttt{metric=neg\_mse} & default \\
        \texttt{max\_orders=100\,000} & \texttt{degree=2}        &        \\
        \texttt{stop\_thresh=1e-10}   & \texttt{max\_terms=2}    &        \\
                                     & \texttt{n\_samples=50\,000} &    \\
                                     & \texttt{batch\_size=1\,000} &   \\
        \bottomrule
    \end{tabular}

    \vspace{0.5em}

    \begin{tabular}{lll}
        \toprule
        E2E\citep{kamienny22_transformer} 
        & Operon\citep{operon_20} 
        & T4SR\citep{lalande23_transformer} \\
        \midrule
        \texttt{beam\_size=20} & default & default \\
                               &         &         \\
                               &         &         \\
        \bottomrule
    \end{tabular}

    \caption{Parameter settings for different symbolic regression algorithms used in our experiments.}
    \label{tab:param_settings}
\end{table}

\subsection{Creating SynEq Dataset}
\label{subsec:creating_syneq}
In Section 3 of the main text, we describe how the development set of the benchmark is constructed. The SynEq dataset in the development set is created by randomly selecting formulas and corresponding intervals that are disjunct from the other formulas in the development set.

\paragraph{Sampling of Formulas.}
In order to generate random formulas, we use the formula generator by \citet{udfs_kahlmeyer24}, which constructs formulas as directed, acyclic graphs (DAGs).
We used this generator to produce formulas with \textbf{at most one constant and at most three input variables}. The constant was set to a random float in the interval $[-10, 10]$. Note, however, that the single constant can branch into multiple constant. For example, if the formula $y = c+x_0\sin(c)$ is instantiated with $c = 1$, the formula becomes $y = 1 +0.841x_0$.

From these formulas, we reject those who are contained in the development set or in the already accepted formulas. To check this condition, we numerically compare the function values at 100 fixed input points, as this can be done efficiently. Although this numerical check may lead to false negatives, these errors are acceptable as they only falsely reject formulas. The accepted formulas are guaranteed to be different.

As discussed in Section 2 of the paper, a benchmark should cover formulas of different complexities. 
We used the number of operations in the corresponding SymPy expression tree (using \texttt{sympy.count\_ops}) as a measure of complexity. We then generated formulas, until we had 200 accepted formulas for the complexities 4-15 and the number of variables 1-3. That is, a total of 12$\times$3$\times$200 = 7\,200 unique and disjunct formulas in total.
In order to avoid formulas that can already be recovered by standard linear regression, we omitted formulas where a standard linear regressor has an $R^2$-Score greater than 0.999.

From these expressions we finally selected 5303 for the development set.
\begin{figure}
    \centering
    \includegraphics[width=\linewidth]{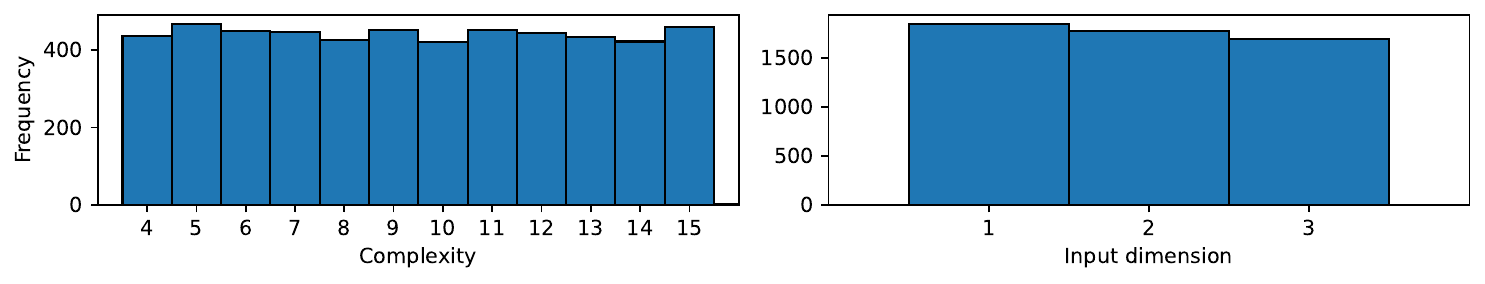}
    \caption{Frequency of complexities and input dimensions in the SynEq dataset. Complexity is measured as the number of operations in the corresponding SymPy expression tree.}
    \label{fig:histograms_syneq}
\end{figure}
Figure~\ref{fig:histograms_syneq} shows that the resulting dataset contains expressions uniformly distributed among the complexities and number of variables (input dimension).

\paragraph{Selecting Intervals.}
In order to evaluate the performance of symbolic regression algorithms, we need samples for the independent and dependent variables.
That is, given the formulas, we need to find a box for the independent variables in which the function can safely be evaluated.
\begin{figure}[t]
    \centering
    \includegraphics[width=\linewidth]{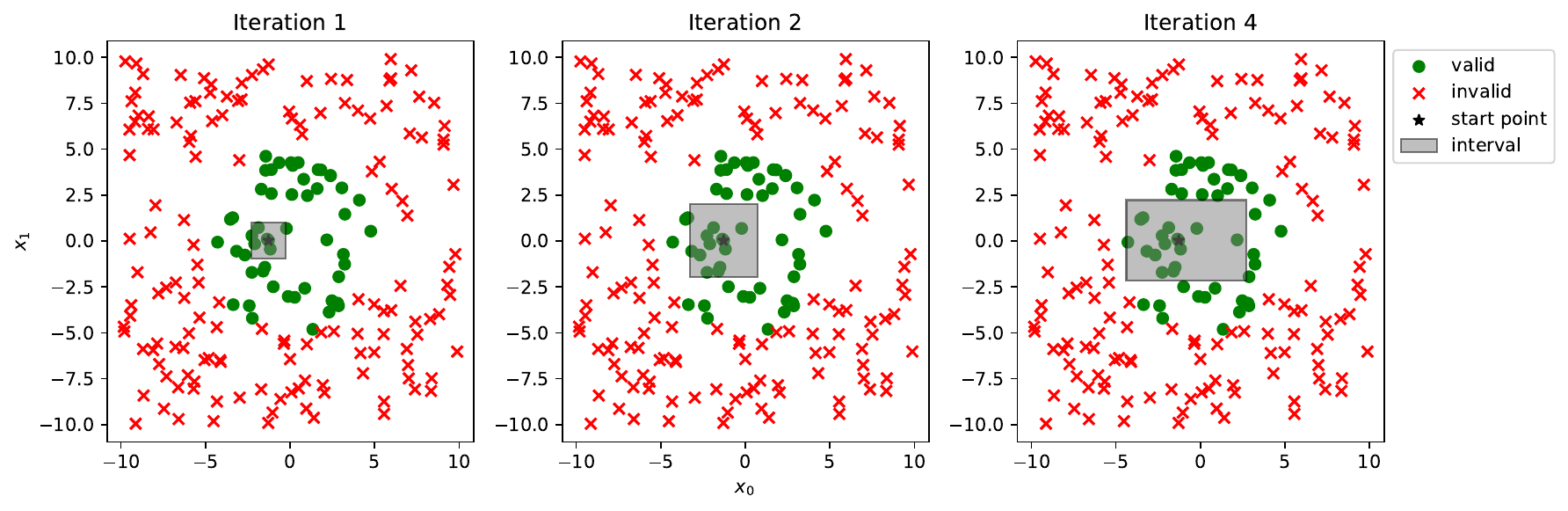}
    \caption{Iterative interval expansion for the equation $y = \log\left(5 - \sqrt{x_0^2 + x_1^2}\right)$. Starting from a valid point, the interval expands gradually in each direction. Finally, we end up with the intervals $x_0\in[-4.47, 4.47], x_1\in[-2.22, 2.22]$.}
    \label{fig:intervals}
\end{figure}
We automate this task of finding a box using the following algorithm, illustrated in Figure~\ref{fig:intervals}:
\begin{enumerate}
    \item Sample from a large box and find one valid point. Set the initial box to the box of zero-width located at this point.
    \item From the current box, gradually expand the space around each dimension. For each expansion, sample from the additional space and check the validity of the samples.
    \item If an expansion fails, the step size is reduced until a minimum step size is reached or if the expansion is valid.
    \item If any dimension has been expanded and the maximum number of expansions is not exceeded, goto 2.
\end{enumerate}
Based upon this algorithm, we were able to generate valid intervals for all the SynEq formulas.

\paragraph{Sampling Points.}
In Section 2 of the paper, we discussed the dependency of pre-training based algorithms on their pre-training distribution. A consequence of this dependence is that a benchmark should not stick to a fixed distribution for sampling.
Therefore, we use a diverse mixture model sampling similar to~\citet{kamienny22_transformer}.
Given a target interval $[a, b]$, we generate samples as follows
\begin{enumerate}
    \item Sample the number of mixture parts $k\sim Cat(1, K)$
    \item Sample the mixture weights $[w_i\sim U(0, 1)]_{i=1,\dots,k}$
    \item Normalize the weight vector $\boldsymbol{w}:= \boldsymbol{w}/||\boldsymbol{w}||$
    \item For each mixture component:
    \begin{enumerate}
        \item Randomly choose a distribution from $[U, \mathcal{N}]$
        \item If uniform, randomly select a sub-interval $[b, c]\subseteq [a, b]$
        \item If normal, randomly select a mean from $\mu~\sim U(a, b)$ and set $\sigma= \min(b-\mu, \mu-a)/3$ 
        \item Generate $\lfloor w_i\cdot N\rfloor$ samples
        \item Clip the samples to $[a, b]$
    \end{enumerate}
    \item Generate the remaining sample points from the first mixture component
\end{enumerate}
In our implementation, we set the maximum number of mixture components to $K=3$.
\begin{figure}[ht]
    \centering
    \includegraphics[width=0.4\linewidth]{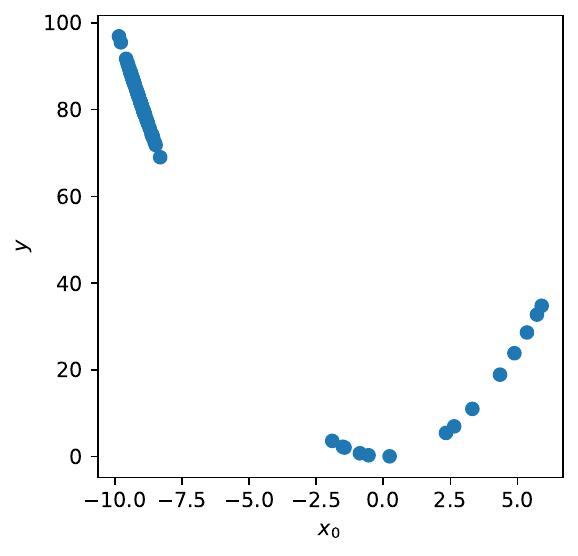}
    \caption{Samples from the diverse sampling strategy for $y = x_0^2$. Points are sampled from a mixture model, where the mixture components are either uniform or gaussian.}
    \label{fig:sampling}
\end{figure}
An example of samples generated for the function $y = x_0^2$ is shown in Figure~\ref{fig:sampling}.

\end{document}

%% file: figures/taxonomy.tex
\tikzset{every picture/.style={line width=0.75pt}} 

\begin{tikzpicture}[x=0.75pt,y=0.75pt,yscale=-1,xscale=1]

\draw  [draw opacity=0][fill={rgb, 255:red, 187; green, 173; blue, 108 }  ,fill opacity=0.5 ] (90,230) -- (130,230) -- (130,250) -- (90,250) -- cycle ;
\draw  [draw opacity=0][fill={rgb, 255:red, 155; green, 155; blue, 155 }  ,fill opacity=0.2 ] (410,177) -- (580,177) -- (580,227) -- (410,227) -- cycle ;
\draw  [draw opacity=0][fill={rgb, 255:red, 155; green, 155; blue, 155 }  ,fill opacity=0.2 ] (220,151) -- (390,151) -- (390,221) -- (220,221) -- cycle ;
\draw  [draw opacity=0][fill={rgb, 255:red, 155; green, 155; blue, 155 }  ,fill opacity=0.2 ] (30,117) -- (200,117) -- (200,223) -- (30,223) -- cycle ;
\draw  [draw opacity=0][fill={rgb, 255:red, 187; green, 173; blue, 108 }  ,fill opacity=0.5 ] (71,172.5) -- (111,172.5) -- (111,200) -- (71,200) -- cycle ;
\draw  [draw opacity=0][fill={rgb, 255:red, 187; green, 173; blue, 108 }  ,fill opacity=0.5 ] (70,127) -- (110,127) -- (110,160) -- (70,160) -- cycle ;
\draw  [draw opacity=0][fill={rgb, 255:red, 124; green, 123; blue, 120 }  ,fill opacity=0.5 ] (290,170) -- (330,170) -- (330,210) -- (290,210) -- cycle ;
\draw    (60,240) -- (165.83,240) ;
\draw [shift={(167.83,240)}, rotate = 180] [color={rgb, 255:red, 0; green, 0; blue, 0 }  ][line width=0.75]    (6.56,-1.97) .. controls (4.17,-0.84) and (1.99,-0.18) .. (0,0) .. controls (1.99,0.18) and (4.17,0.84) .. (6.56,1.97)   ;
\draw    (70,233) -- (70,247) ;
\draw    (110,233) -- (110,247) ;
\draw    (150,233) -- (150,247) ;
\draw    (53,152) -- (121,152) ;
\draw [shift={(123,152)}, rotate = 180] [color={rgb, 255:red, 0; green, 0; blue, 0 }  ][line width=0.75]    (6.56,-1.97) .. controls (4.17,-0.84) and (1.99,-0.18) .. (0,0) .. controls (1.99,0.18) and (4.17,0.84) .. (6.56,1.97)   ;
\draw  [fill={rgb, 255:red, 0; green, 34; blue, 77 }  ,fill opacity=1 ] (88,152) .. controls (88,150.9) and (88.9,150) .. (90,150) .. controls (91.1,150) and (92,150.9) .. (92,152) .. controls (92,153.1) and (91.1,154) .. (90,154) .. controls (88.9,154) and (88,153.1) .. (88,152) -- cycle ;
\draw  [fill={rgb, 255:red, 0; green, 34; blue, 77 }  ,fill opacity=1 ] (95,145) .. controls (95,143.9) and (95.9,143) .. (97,143) .. controls (98.1,143) and (99,143.9) .. (99,145) .. controls (99,146.1) and (98.1,147) .. (97,147) .. controls (95.9,147) and (95,146.1) .. (95,145) -- cycle ;
\draw  [fill={rgb, 255:red, 0; green, 34; blue, 77 }  ,fill opacity=1 ] (103,133) .. controls (103,131.9) and (103.9,131) .. (105,131) .. controls (106.1,131) and (107,131.9) .. (107,133) .. controls (107,134.1) and (106.1,135) .. (105,135) .. controls (103.9,135) and (103,134.1) .. (103,133) -- cycle ;
\draw  [fill={rgb, 255:red, 0; green, 34; blue, 77 }  ,fill opacity=1 ] (80,146) .. controls (80,144.9) and (80.9,144) .. (82,144) .. controls (83.1,144) and (84,144.9) .. (84,146) .. controls (84,147.1) and (83.1,148) .. (82,148) .. controls (80.9,148) and (80,147.1) .. (80,146) -- cycle ;
\draw  [fill={rgb, 255:red, 0; green, 34; blue, 77 }  ,fill opacity=1 ] (70,133) .. controls (70,131.9) and (70.9,131) .. (72,131) .. controls (73.1,131) and (74,131.9) .. (74,133) .. controls (74,134.1) and (73.1,135) .. (72,135) .. controls (70.9,135) and (70,134.1) .. (70,133) -- cycle ;
\draw    (50,192) -- (118,192) ;
\draw [shift={(120,192)}, rotate = 180] [color={rgb, 255:red, 0; green, 0; blue, 0 }  ][line width=0.75]    (6.56,-1.97) .. controls (4.17,-0.84) and (1.99,-0.18) .. (0,0) .. controls (1.99,0.18) and (4.17,0.84) .. (6.56,1.97)   ;
\draw  [fill={rgb, 255:red, 0; green, 34; blue, 77 }  ,fill opacity=1 ] (88,190) .. controls (88,188.9) and (88.9,188) .. (90,188) .. controls (91.1,188) and (92,188.9) .. (92,190) .. controls (92,191.1) and (91.1,192) .. (90,192) .. controls (88.9,192) and (88,191.1) .. (88,190) -- cycle ;
\draw  [fill={rgb, 255:red, 0; green, 34; blue, 77 }  ,fill opacity=1 ] (97,181) .. controls (97,179.9) and (97.9,179) .. (99,179) .. controls (100.1,179) and (101,179.9) .. (101,181) .. controls (101,182.1) and (100.1,183) .. (99,183) .. controls (97.9,183) and (97,182.1) .. (97,181) -- cycle ;
\draw  [fill={rgb, 255:red, 0; green, 34; blue, 77 }  ,fill opacity=1 ] (106,179) .. controls (106,177.9) and (106.9,177) .. (108,177) .. controls (109.1,177) and (110,177.9) .. (110,179) .. controls (110,180.1) and (109.1,181) .. (108,181) .. controls (106.9,181) and (106,180.1) .. (106,179) -- cycle ;
\draw  [fill={rgb, 255:red, 0; green, 34; blue, 77 }  ,fill opacity=1 ] (80,197) .. controls (80,195.9) and (80.9,195) .. (82,195) .. controls (83.1,195) and (84,195.9) .. (84,197) .. controls (84,198.1) and (83.1,199) .. (82,199) .. controls (80.9,199) and (80,198.1) .. (80,197) -- cycle ;
\draw  [fill={rgb, 255:red, 0; green, 34; blue, 77 }  ,fill opacity=1 ] (70,200) .. controls (70,198.9) and (70.9,198) .. (72,198) .. controls (73.1,198) and (74,198.9) .. (74,200) .. controls (74,201.1) and (73.1,202) .. (72,202) .. controls (70.9,202) and (70,201.1) .. (70,200) -- cycle ;
\draw    (130,120) -- (130,220) ;
\draw    (50,160) -- (180,160) ;
\draw    (50,202) -- (180,202) ;
\draw    (50,170) -- (40,170) -- (40,100) -- (48,100) ;
\draw [shift={(50,100)}, rotate = 180] [fill={rgb, 255:red, 0; green, 0; blue, 0 }  ][line width=0.08]  [draw opacity=0] (7.2,-1.8) -- (0,0) -- (7.2,1.8) -- cycle    ;
\draw    (180,171) -- (190,171) -- (190,100) -- (182,100) ;
\draw [shift={(180,100)}, rotate = 360] [fill={rgb, 255:red, 0; green, 0; blue, 0 }  ][line width=0.08]  [draw opacity=0] (7.2,-1.8) -- (0,0) -- (7.2,1.8) -- cycle    ;
\draw   (50,94.5) .. controls (50,92.01) and (52.01,90) .. (54.5,90) -- (175.5,90) .. controls (177.99,90) and (180,92.01) .. (180,94.5) -- (180,108) .. controls (180,110.49) and (177.99,112.5) .. (175.5,112.5) -- (54.5,112.5) .. controls (52.01,112.5) and (50,110.49) .. (50,108) -- cycle ;
\draw    (270,200) -- (338,200) ;
\draw [shift={(340,200)}, rotate = 180] [color={rgb, 255:red, 0; green, 0; blue, 0 }  ][line width=0.75]    (6.56,-1.97) .. controls (4.17,-0.84) and (1.99,-0.18) .. (0,0) .. controls (1.99,0.18) and (4.17,0.84) .. (6.56,1.97)   ;
\draw  [fill={rgb, 255:red, 0; green, 34; blue, 77 }  ,fill opacity=1 ] (308,184) .. controls (308,182.9) and (308.9,182) .. (310,182) .. controls (311.1,182) and (312,182.9) .. (312,184) .. controls (312,185.1) and (311.1,186) .. (310,186) .. controls (308.9,186) and (308,185.1) .. (308,184) -- cycle ;
\draw  [fill={rgb, 255:red, 0; green, 34; blue, 77 }  ,fill opacity=1 ] (315,181) .. controls (315,179.9) and (315.9,179) .. (317,179) .. controls (318.1,179) and (319,179.9) .. (319,181) .. controls (319,182.1) and (318.1,183) .. (317,183) .. controls (315.9,183) and (315,182.1) .. (315,181) -- cycle ;
\draw  [fill={rgb, 255:red, 0; green, 34; blue, 77 }  ,fill opacity=1 ] (323,179) .. controls (323,177.9) and (323.9,177) .. (325,177) .. controls (326.1,177) and (327,177.9) .. (327,179) .. controls (327,180.1) and (326.1,181) .. (325,181) .. controls (323.9,181) and (323,180.1) .. (323,179) -- cycle ;
\draw  [fill={rgb, 255:red, 0; green, 34; blue, 77 }  ,fill opacity=1 ] (296,197) .. controls (296,195.9) and (296.9,195) .. (298,195) .. controls (299.1,195) and (300,195.9) .. (300,197) .. controls (300,198.1) and (299.1,199) .. (298,199) .. controls (296.9,199) and (296,198.1) .. (296,197) -- cycle ;
\draw  [fill={rgb, 255:red, 0; green, 34; blue, 77 }  ,fill opacity=1 ] (300,189) .. controls (300,187.9) and (300.9,187) .. (302,187) .. controls (303.1,187) and (304,187.9) .. (304,189) .. controls (304,190.1) and (303.1,191) .. (302,191) .. controls (300.9,191) and (300,190.1) .. (300,189) -- cycle ;
\draw   (230,123.35) .. controls (230,121.09) and (231.84,119.25) .. (234.1,119.25) -- (375.9,119.25) .. controls (378.16,119.25) and (380,121.09) .. (380,123.35) -- (380,135.65) .. controls (380,137.91) and (378.16,139.75) .. (375.9,139.75) -- (234.1,139.75) .. controls (231.84,139.75) and (230,137.91) .. (230,135.65) -- cycle ;
\draw   (420,148.6) .. controls (420,146.06) and (422.06,144) .. (424.6,144) -- (565.4,144) .. controls (567.94,144) and (570,146.06) .. (570,148.6) -- (570,162.4) .. controls (570,164.94) and (567.94,167) .. (565.4,167) -- (424.6,167) .. controls (422.06,167) and (420,164.94) .. (420,162.4) -- cycle ;
\draw    (490,145) -- (490,132) ;
\draw [shift={(490,130)}, rotate = 90] [fill={rgb, 255:red, 0; green, 0; blue, 0 }  ][line width=0.08]  [draw opacity=0] (7.2,-1.8) -- (0,0) -- (7.2,1.8) -- cycle    ;
\draw    (490,182) -- (490,169) ;
\draw [shift={(490,167)}, rotate = 90] [fill={rgb, 255:red, 0; green, 0; blue, 0 }  ][line width=0.08]  [draw opacity=0] (7.2,-1.8) -- (0,0) -- (7.2,1.8) -- cycle    ;
\draw  [draw opacity=0][fill={rgb, 255:red, 67; green, 78; blue, 107 }  ,fill opacity=0.5 ] (470,90) -- (510,90) -- (510,130) -- (470,130) -- cycle ;
\draw    (450,120) -- (528,120) ;
\draw [shift={(530,120)}, rotate = 180] [color={rgb, 255:red, 0; green, 0; blue, 0 }  ][line width=0.75]    (6.56,-1.97) .. controls (4.17,-0.84) and (1.99,-0.18) .. (0,0) .. controls (1.99,0.18) and (4.17,0.84) .. (6.56,1.97)   ;
\draw  [fill={rgb, 255:red, 0; green, 34; blue, 77 }  ,fill opacity=1 ] (488,108) .. controls (488,106.9) and (488.9,106) .. (490,106) .. controls (491.1,106) and (492,106.9) .. (492,108) .. controls (492,109.1) and (491.1,110) .. (490,110) .. controls (488.9,110) and (488,109.1) .. (488,108) -- cycle ;
\draw  [fill={rgb, 255:red, 0; green, 34; blue, 77 }  ,fill opacity=1 ] (493,112) .. controls (493,110.9) and (493.9,110) .. (495,110) .. controls (496.1,110) and (497,110.9) .. (497,112) .. controls (497,113.1) and (496.1,114) .. (495,114) .. controls (493.9,114) and (493,113.1) .. (493,112) -- cycle ;
\draw  [fill={rgb, 255:red, 0; green, 34; blue, 77 }  ,fill opacity=1 ] (503,125) .. controls (503,123.9) and (503.9,123) .. (505,123) .. controls (506.1,123) and (507,123.9) .. (507,125) .. controls (507,126.1) and (506.1,127) .. (505,127) .. controls (503.9,127) and (503,126.1) .. (503,125) -- cycle ;
\draw  [fill={rgb, 255:red, 0; green, 34; blue, 77 }  ,fill opacity=1 ] (483,105) .. controls (483,103.9) and (483.9,103) .. (485,103) .. controls (486.1,103) and (487,103.9) .. (487,105) .. controls (487,106.1) and (486.1,107) .. (485,107) .. controls (483.9,107) and (483,106.1) .. (483,105) -- cycle ;
\draw  [fill={rgb, 255:red, 0; green, 34; blue, 77 }  ,fill opacity=1 ] (470,101) .. controls (470,99.9) and (470.9,99) .. (472,99) .. controls (473.1,99) and (474,99.9) .. (474,101) .. controls (474,102.1) and (473.1,103) .. (472,103) .. controls (470.9,103) and (470,102.1) .. (470,101) -- cycle ;
\draw  [draw opacity=0][fill={rgb, 255:red, 67; green, 78; blue, 107 }  ,fill opacity=0.5 ] (480,190) -- (520,190) -- (520,210) -- (480,210) -- cycle ;
\draw    (460,200) -- (528,200) ;
\draw [shift={(530,200)}, rotate = 180] [color={rgb, 255:red, 0; green, 0; blue, 0 }  ][line width=0.75]    (6.56,-1.97) .. controls (4.17,-0.84) and (1.99,-0.18) .. (0,0) .. controls (1.99,0.18) and (4.17,0.84) .. (6.56,1.97)   ;
\draw  [fill={rgb, 255:red, 0; green, 34; blue, 77 }  ,fill opacity=1 ] (498,200) .. controls (498,198.9) and (498.9,198) .. (500,198) .. controls (501.1,198) and (502,198.9) .. (502,200) .. controls (502,201.1) and (501.1,202) .. (500,202) .. controls (498.9,202) and (498,201.1) .. (498,200) -- cycle ;
\draw  [fill={rgb, 255:red, 0; green, 34; blue, 77 }  ,fill opacity=1 ] (503,200) .. controls (503,198.9) and (503.9,198) .. (505,198) .. controls (506.1,198) and (507,198.9) .. (507,200) .. controls (507,201.1) and (506.1,202) .. (505,202) .. controls (503.9,202) and (503,201.1) .. (503,200) -- cycle ;
\draw  [fill={rgb, 255:red, 0; green, 34; blue, 77 }  ,fill opacity=1 ] (513,200) .. controls (513,198.9) and (513.9,198) .. (515,198) .. controls (516.1,198) and (517,198.9) .. (517,200) .. controls (517,201.1) and (516.1,202) .. (515,202) .. controls (513.9,202) and (513,201.1) .. (513,200) -- cycle ;
\draw  [fill={rgb, 255:red, 0; green, 34; blue, 77 }  ,fill opacity=1 ] (493,200) .. controls (493,198.9) and (493.9,198) .. (495,198) .. controls (496.1,198) and (497,198.9) .. (497,200) .. controls (497,201.1) and (496.1,202) .. (495,202) .. controls (493.9,202) and (493,201.1) .. (493,200) -- cycle ;
\draw  [fill={rgb, 255:red, 0; green, 34; blue, 77 }  ,fill opacity=1 ] (480,200) .. controls (480,198.9) and (480.9,198) .. (482,198) .. controls (483.1,198) and (484,198.9) .. (484,200) .. controls (484,201.1) and (483.1,202) .. (482,202) .. controls (480.9,202) and (480,201.1) .. (480,200) -- cycle ;
\draw    (460,130) -- (460,82) ;
\draw [shift={(460,80)}, rotate = 90] [color={rgb, 255:red, 0; green, 0; blue, 0 }  ][line width=0.75]    (6.56,-1.97) .. controls (4.17,-0.84) and (1.99,-0.18) .. (0,0) .. controls (1.99,0.18) and (4.17,0.84) .. (6.56,1.97)   ;
\draw    (290,210) -- (290,162) ;
\draw [shift={(290,160)}, rotate = 90] [color={rgb, 255:red, 0; green, 0; blue, 0 }  ][line width=0.75]    (6.56,-1.97) .. controls (4.17,-0.84) and (1.99,-0.18) .. (0,0) .. controls (1.99,0.18) and (4.17,0.84) .. (6.56,1.97)   ;
\draw    (60,160) -- (60,132) ;
\draw [shift={(60,130)}, rotate = 90] [color={rgb, 255:red, 0; green, 0; blue, 0 }  ][line width=0.75]    (6.56,-1.97) .. controls (4.17,-0.84) and (1.99,-0.18) .. (0,0) .. controls (1.99,0.18) and (4.17,0.84) .. (6.56,1.97)   ;
\draw    (60,202) -- (60,174) ;
\draw [shift={(60,172)}, rotate = 90] [color={rgb, 255:red, 0; green, 0; blue, 0 }  ][line width=0.75]    (6.56,-1.97) .. controls (4.17,-0.84) and (1.99,-0.18) .. (0,0) .. controls (1.99,0.18) and (4.17,0.84) .. (6.56,1.97)   ;
\draw   (30,94.17) .. controls (30,86.34) and (36.34,80) .. (44.17,80) -- (185.83,80) .. controls (193.66,80) and (200,86.34) .. (200,94.17) -- (200,255.83) .. controls (200,263.66) and (193.66,270) .. (185.83,270) -- (44.17,270) .. controls (36.34,270) and (30,263.66) .. (30,255.83) -- cycle ;
\draw   (50,120) -- (180,120) -- (180,220) -- (50,220) -- cycle ;
\draw   (220,94.17) .. controls (220,86.34) and (226.34,80) .. (234.17,80) -- (375.83,80) .. controls (383.66,80) and (390,86.34) .. (390,94.17) -- (390,255.83) .. controls (390,263.66) and (383.66,270) .. (375.83,270) -- (234.17,270) .. controls (226.34,270) and (220,263.66) .. (220,255.83) -- cycle ;
\draw   (410,94.17) .. controls (410,86.34) and (416.34,80) .. (424.17,80) -- (565.83,80) .. controls (573.66,80) and (580,86.34) .. (580,94.17) -- (580,255.83) .. controls (580,263.66) and (573.66,270) .. (565.83,270) -- (424.17,270) .. controls (416.34,270) and (410,263.66) .. (410,255.83) -- cycle ;
\draw    (306,160) -- (306,142) ;
\draw [shift={(306,140)}, rotate = 90] [fill={rgb, 255:red, 0; green, 0; blue, 0 }  ][line width=0.08]  [draw opacity=0] (7.2,-1.8) -- (0,0) -- (7.2,1.8) -- cycle    ;
\draw    (306,120) -- (306,105) ;
\draw [shift={(306,103)}, rotate = 90] [fill={rgb, 255:red, 0; green, 0; blue, 0 }  ][line width=0.08]  [draw opacity=0] (7.2,-1.8) -- (0,0) -- (7.2,1.8) -- cycle    ;
\draw  [dash pattern={on 0.84pt off 2.51pt}]  (110,90) -- (110,70) -- (210,70) -- (210,130) -- (228,130) ;
\draw [shift={(230,130)}, rotate = 180] [fill={rgb, 255:red, 0; green, 0; blue, 0 }  ][line width=0.08]  [draw opacity=0] (7.2,-1.8) -- (0,0) -- (7.2,1.8) -- cycle    ;
\draw  [dash pattern={on 0.84pt off 2.51pt}]  (305,86) -- (305,70) -- (400,70) -- (400,155) -- (418,155) ;
\draw [shift={(420,155)}, rotate = 180] [fill={rgb, 255:red, 0; green, 0; blue, 0 }  ][line width=0.08]  [draw opacity=0] (7.2,-1.8) -- (0,0) -- (7.2,1.8) -- cycle    ;
\draw    (90,230) -- (90,250) ;
\draw    (130,230) -- (130,250) ;
\draw  [draw opacity=0][fill={rgb, 255:red, 124; green, 123; blue, 120 }  ,fill opacity=0.5 ] (260,230) -- (300,230) -- (300,250) -- (260,250) -- cycle ;
\draw  [draw opacity=0][fill={rgb, 255:red, 67; green, 78; blue, 107 }  ,fill opacity=0.5 ] (490,230) -- (530,230) -- (530,250) -- (490,250) -- cycle ;
\draw    (250,240) -- (355.83,240) ;
\draw [shift={(357.83,240)}, rotate = 180] [color={rgb, 255:red, 0; green, 0; blue, 0 }  ][line width=0.75]    (6.56,-1.97) .. controls (4.17,-0.84) and (1.99,-0.18) .. (0,0) .. controls (1.99,0.18) and (4.17,0.84) .. (6.56,1.97)   ;
\draw    (340,233) -- (340,247) ;
\draw    (260,230) -- (260,250) ;
\draw    (300,230) -- (300,250) ;
\draw    (440,240) -- (545.83,240) ;
\draw [shift={(547.83,240)}, rotate = 180] [color={rgb, 255:red, 0; green, 0; blue, 0 }  ][line width=0.75]    (6.56,-1.97) .. controls (4.17,-0.84) and (1.99,-0.18) .. (0,0) .. controls (1.99,0.18) and (4.17,0.84) .. (6.56,1.97)   ;
\draw    (450,233) -- (450,247) ;
\draw    (490,230) -- (490,250) ;
\draw    (530,230) -- (530,250) ;

\draw (160,242.4) node [anchor=north west][inner sep=0.75pt]    {$x$};
\draw (107,252.4) node [anchor=north west][inner sep=0.75pt]    {$0$};
\draw (51,252.4) node [anchor=north west][inner sep=0.75pt]    {$-1$};
\draw (147,252.4) node [anchor=north west][inner sep=0.75pt]    {$1$};
\draw (140,132.4) node [anchor=north west][inner sep=0.75pt]    {$x^{2}$};
\draw (132,172.4) node [anchor=north west][inner sep=0.75pt]    {$\sin( x)$};
\draw (115,101.25) node   [align=left] {Model};
\draw (305,129.5) node   [align=left] {Regression Algorithm};
\draw (284,82.4) node [anchor=north west][inner sep=0.75pt]    {$\cos( x)$};
\draw (495,155.5) node   [align=left] {Regression Function};
\draw (113,271) node [anchor=north] [inner sep=0.75pt]   [align=left] {Pre-Training};
\draw (301.5,271) node [anchor=north] [inner sep=0.75pt]   [align=left] {Training};
\draw (493,271) node [anchor=north] [inner sep=0.75pt]   [align=left] {Inference};
\draw (74,216.6) node [anchor=south west] [inner sep=0.75pt]    {$\dotsc $};
\draw (164,216.6) node [anchor=south east] [inner sep=0.75pt]    {$\dotsc $};
\draw (11,274) node [anchor=north west][inner sep=0.75pt]  [rotate=-270] [align=left] {Domain};
\draw (11,114) node [anchor=north west][inner sep=0.75pt]  [rotate=-270] [align=left] {Task};
\draw (11,189) node [anchor=north west][inner sep=0.75pt]  [rotate=-270] [align=left] {Data};
\draw (350,242.4) node [anchor=north west][inner sep=0.75pt]    {$x$};
\draw (297,252.4) node [anchor=north west][inner sep=0.75pt]    {$0$};
\draw (241,252.4) node [anchor=north west][inner sep=0.75pt]    {$-1$};
\draw (337,252.4) node [anchor=north west][inner sep=0.75pt]    {$1$};
\draw (540,242.4) node [anchor=north west][inner sep=0.75pt]    {$x$};
\draw (487,252.4) node [anchor=north west][inner sep=0.75pt]    {$0$};
\draw (431,252.4) node [anchor=north west][inner sep=0.75pt]    {$-1$};
\draw (527,252.4) node [anchor=north west][inner sep=0.75pt]    {$1$};

\end{tikzpicture}

%% file: figures/eval_protocol1.tex
\tikzset{every picture/.style={line width=0.75pt}} 

\begin{tikzpicture}[x=0.75pt,y=0.75pt,yscale=-1,xscale=1]

\draw  [fill={rgb, 255:red, 255; green, 255; blue, 255 }  ,fill opacity=1 ][general shadow={fill={rgb, 255:red, 155; green, 155; blue, 155 }  ,shadow xshift=0.75pt,shadow yshift=-0.75pt, opacity=1 }] (30,33.4) .. controls (30,31.25) and (31.75,29.5) .. (33.9,29.5) -- (196.1,29.5) .. controls (198.25,29.5) and (200,31.25) .. (200,33.4) -- (200,45.1) .. controls (200,47.25) and (198.25,49) .. (196.1,49) -- (33.9,49) .. controls (31.75,49) and (30,47.25) .. (30,45.1) -- cycle ;
\draw  [fill={rgb, 255:red, 255; green, 255; blue, 255 }  ,fill opacity=1 ][general shadow={fill={rgb, 255:red, 155; green, 155; blue, 155 }  ,shadow xshift=0.75pt,shadow yshift=-0.75pt, opacity=1 }] (30,84) .. controls (30,81.79) and (31.79,80) .. (34,80) -- (196,80) .. controls (198.21,80) and (200,81.79) .. (200,84) -- (200,96) .. controls (200,98.21) and (198.21,100) .. (196,100) -- (34,100) .. controls (31.79,100) and (30,98.21) .. (30,96) -- cycle ;
\draw    (190,80) -- (190,52) ;
\draw [shift={(190,50)}, rotate = 90] [fill={rgb, 255:red, 0; green, 0; blue, 0 }  ][line width=0.08]  [draw opacity=0] (12,-3) -- (0,0) -- (12,3) -- cycle    ;
\draw    (190,150) -- (190,102) ;
\draw [shift={(190,100)}, rotate = 90] [fill={rgb, 255:red, 0; green, 0; blue, 0 }  ][line width=0.08]  [draw opacity=0] (12,-3) -- (0,0) -- (12,3) -- cycle    ;
\draw    (40,100) .. controls (42.15,150.99) and (102.93,159.34) .. (148.62,159.98) ;
\draw [shift={(150,160)}, rotate = 180.63] [fill={rgb, 255:red, 0; green, 0; blue, 0 }  ][line width=0.08]  [draw opacity=0] (12,-3) -- (0,0) -- (12,3) -- cycle    ;
\draw    (40,100) .. controls (47.02,156.35) and (154.73,120.02) .. (178.64,101.13) ;
\draw [shift={(180,100)}, rotate = 138.39] [fill={rgb, 255:red, 0; green, 0; blue, 0 }  ][line width=0.08]  [draw opacity=0] (12,-3) -- (0,0) -- (12,3) -- cycle    ;
\draw    (40,50) -- (40,78.5) ;
\draw [shift={(40,80.5)}, rotate = 270] [fill={rgb, 255:red, 0; green, 0; blue, 0 }  ][line width=0.08]  [draw opacity=0] (12,-3) -- (0,0) -- (12,3) -- cycle    ;
\draw    (110,170) -- (148,170) ;
\draw [shift={(150,170)}, rotate = 180] [fill={rgb, 255:red, 0; green, 0; blue, 0 }  ][line width=0.08]  [draw opacity=0] (12,-3) -- (0,0) -- (12,3) -- cycle    ;

\draw (110,39.5) node   [align=left] {User};
\draw (115,90) node   [align=left] {Competition Website};
\draw (22,102) node [anchor=north west][inner sep=0.75pt]   [align=left] {ID};
\draw (42,65.25) node [anchor=west] [inner sep=0.75pt]   [align=left] {Request};
\draw (192,65) node [anchor=west] [inner sep=0.75pt]   [align=left] {ID, $\displaystyle \left(\tilde{X} ,\tilde{\boldsymbol{y}}\right)$};
\draw (21,181) node [anchor=north west][inner sep=0.75pt]   [align=left] {Evaluation Set};
\draw (27,161.4) node [anchor=north west][inner sep=0.75pt]    {$( X,\ \boldsymbol{y} ,\ f)$};
\draw (151,181) node [anchor=north west][inner sep=0.75pt]   [align=left] {Permutation};
\draw (151,157.4) node [anchor=north west][inner sep=0.75pt]    {$\left(\tilde{X} ,\ \tilde{\boldsymbol{y}} ,\ \tilde{f}\right)$};

\end{tikzpicture}

%% file: figures/eval_protocol2.tex
\tikzset{every picture/.style={line width=0.75pt}} 

\begin{tikzpicture}[x=0.75pt,y=0.75pt,yscale=-1,xscale=1]

\draw  [fill={rgb, 255:red, 255; green, 255; blue, 255 }  ,fill opacity=1 ][general shadow={fill={rgb, 255:red, 155; green, 155; blue, 155 }  ,shadow xshift=0.75pt,shadow yshift=-0.75pt, opacity=1 }] (170,114.5) .. controls (170,112.29) and (171.79,110.5) .. (174,110.5) -- (226,110.5) .. controls (228.21,110.5) and (230,112.29) .. (230,114.5) -- (230,126.5) .. controls (230,128.71) and (228.21,130.5) .. (226,130.5) -- (174,130.5) .. controls (171.79,130.5) and (170,128.71) .. (170,126.5) -- cycle ;
\draw  [fill={rgb, 255:red, 255; green, 255; blue, 255 }  ,fill opacity=1 ][general shadow={fill={rgb, 255:red, 155; green, 155; blue, 155 }  ,shadow xshift=0.75pt,shadow yshift=-0.75pt, opacity=1 }] (40,75.2) .. controls (40,72.88) and (41.88,71) .. (44.2,71) -- (225.8,71) .. controls (228.12,71) and (230,72.88) .. (230,75.2) -- (230,87.8) .. controls (230,90.12) and (228.12,92) .. (225.8,92) -- (44.2,92) .. controls (41.88,92) and (40,90.12) .. (40,87.8) -- cycle ;
\draw  [fill={rgb, 255:red, 255; green, 255; blue, 255 }  ,fill opacity=1 ][general shadow={fill={rgb, 255:red, 155; green, 155; blue, 155 }  ,shadow xshift=0.75pt,shadow yshift=-0.75pt, opacity=1 }] (40,24.4) .. controls (40,22.25) and (41.75,20.5) .. (43.9,20.5) -- (216.1,20.5) .. controls (218.25,20.5) and (220,22.25) .. (220,24.4) -- (220,36.1) .. controls (220,38.25) and (218.25,40) .. (216.1,40) -- (43.9,40) .. controls (41.75,40) and (40,38.25) .. (40,36.1) -- cycle ;
\draw    (50,40) -- (50,69.5) ;
\draw [shift={(50,71.5)}, rotate = 270] [fill={rgb, 255:red, 0; green, 0; blue, 0 }  ][line width=0.08]  [draw opacity=0] (12,-3) -- (0,0) -- (12,3) -- cycle    ;
\draw    (120,162) -- (158,162) ;
\draw [shift={(160,162)}, rotate = 180] [fill={rgb, 255:red, 0; green, 0; blue, 0 }  ][line width=0.08]  [draw opacity=0] (12,-3) -- (0,0) -- (12,3) -- cycle    ;
\draw    (200,71) -- (200,42) ;
\draw [shift={(200,40)}, rotate = 90] [fill={rgb, 255:red, 0; green, 0; blue, 0 }  ][line width=0.08]  [draw opacity=0] (12,-3) -- (0,0) -- (12,3) -- cycle    ;
\draw    (50,92) .. controls (52.15,142.99) and (112.93,151.34) .. (158.62,151.98) ;
\draw [shift={(160,152)}, rotate = 180.63] [fill={rgb, 255:red, 0; green, 0; blue, 0 }  ][line width=0.08]  [draw opacity=0] (12,-3) -- (0,0) -- (12,3) -- cycle    ;
\draw    (50,92) .. controls (64.79,123.69) and (141.21,121.63) .. (168.04,120.11) ;
\draw [shift={(170,120)}, rotate = 176.54] [fill={rgb, 255:red, 0; green, 0; blue, 0 }  ][line width=0.08]  [draw opacity=0] (12,-3) -- (0,0) -- (12,3) -- cycle    ;
\draw    (200,150) -- (200,132) ;
\draw [shift={(200,130)}, rotate = 90] [fill={rgb, 255:red, 0; green, 0; blue, 0 }  ][line width=0.08]  [draw opacity=0] (12,-3) -- (0,0) -- (12,3) -- cycle    ;
\draw    (200,110) -- (200,93) ;
\draw [shift={(200,91)}, rotate = 90] [fill={rgb, 255:red, 0; green, 0; blue, 0 }  ][line width=0.08]  [draw opacity=0] (12,-3) -- (0,0) -- (12,3) -- cycle    ;

\draw (130,30.25) node   [align=left] {User};
\draw (130,81.5) node   [align=left] {Competition Website};
\draw (35,172) node [anchor=north west][inner sep=0.75pt]   [align=left] {Evaluation Set};
\draw (123,134) node [anchor=north west][inner sep=0.75pt]   [align=left] {ID};
\draw (41,152.4) node [anchor=north west][inner sep=0.75pt]    {$( X,\ \boldsymbol{y} ,\ f)$};
\draw (161,172) node [anchor=north west][inner sep=0.75pt]   [align=left] {Permutation};
\draw (161,148.4) node [anchor=north west][inner sep=0.75pt]    {$\left(\tilde{X} ,\ \tilde{\boldsymbol{y}} ,\ \tilde{f}\right)$};
\draw (52,56.25) node [anchor=west] [inner sep=0.75pt]   [align=left] {ID, $\displaystyle \tilde{g}$};
\draw (202,55.5) node [anchor=west] [inner sep=0.75pt]   [align=left] {\begin{minipage}[lt]{49.82pt}\setlength\topsep{0pt}
\begin{center}
Scores
\end{center}

\end{minipage}};
\draw (200,120.5) node   [align=left] {Metrics};
\draw (118,102.4) node [anchor=north west][inner sep=0.75pt]    {$\tilde{g}$};

\end{tikzpicture}